\documentclass[10pt,twocolumn,letterpaper]{article}

\usepackage[pagenumbers]{cvpr} 

\makeatletter\let\captiontemp\@makecaption\makeatother
\usepackage[font=small]{subcaption}
\makeatletter\let\@makecaption\captiontemp\makeatother
\usepackage{graphicx}
\usepackage{tabularx}
\usepackage{amsmath}
\usepackage{amssymb}
\usepackage{booktabs}
\usepackage{sg-ibs-macros} 
 \usepackage{array, diagbox}
 \usepackage{pifont}
 \usepackage{stackengine}
\newcommand{\cmark}{\ding{51}}
\newcommand{\xmark}{\ding{55}}

\newcolumntype{M}[1]{>{\centering\arraybackslash}m{#1}}

\definecolor{darkred}{rgb}{0.7,0.1,0.1}
\definecolor{darkgreen}{rgb}{0.1,0.7,0.1}
\definecolor{dblue}{rgb}{0.2,0.2,0.8}
\definecolor{maroon}{rgb}{0.76,.13,.28}
\definecolor{burntorange}{rgb}{0.81,.33,0}
\definecolor{cyan}{rgb}{0.0,0.7,0.94}
\definecolor{salmon}{rgb}{0.99,0.51,0.46}
\definecolor{green}{rgb}{0.03,0.91,0.43}

\definecolor{tpatch_blue}{RGB}{68, 67, 118}
\definecolor{tpatch_pink}{RGB}{119, 56, 75}
\definecolor{tpatch_cyan}{RGB}{22, 119, 131}
\definecolor{tpatch_green}{RGB}{36, 130, 92}

\ifdefined\ShowNotes
  \newcommand{\colornote}[3]{{\color{#1}\bf{#2: #3}\normalfont}}
\else
  \newcommand{\colornote}[3]{}
\fi

\usepackage[pagebackref,breaklinks,colorlinks,citecolor=cvprblue]{hyperref}
\usepackage[capitalize]{cleveref}
\crefname{section}{Sec.}{Secs.}
\Crefname{section}{Section}{Sections}
\Crefname{table}{Table}{Tables}
\crefname{table}{Tab.}{Tabs.}

\definecolor{cvprblue}{rgb}{0.21,0.49,0.74}

\def\paperID{2056} 
\def\confName{CVPR}
\def\confYear{2024}

\usepackage{overpic}
\usepackage{enumitem} 
\usepackage{overpic} 
\usepackage{color}
\usepackage{multirow}

\definecolor{turquoise}{cmyk}{0.65,0,0.1,0.3}
\definecolor{purple}{rgb}{0.65,0,0.65}
\definecolor{dark_green}{rgb}{0, 0.5, 0}
\definecolor{orange}{rgb}{0.8, 0.6, 0.2}
\definecolor{red}{rgb}{0.8, 0.2, 0.2}
\definecolor{darkred}{rgb}{0.6, 0.1, 0.05}
\definecolor{blueish}{rgb}{0.0, 0.3, .6}
\definecolor{light_gray}{rgb}{0.7, 0.7, .7}
\definecolor{pink}{rgb}{1, 0, 1}
\definecolor{greyblue}{rgb}{0.25, 0.25, 1}

\usepackage{blindtext}

\renewcommand{\paragraph}[1]{\vspace{1em}\noindent\textbf{#1}.}
\begin{document}

\teaser{
    \centering
    \includegraphics[width=\linewidth,trim={1cm 5.5cm 1cm 4.5cm},clip]{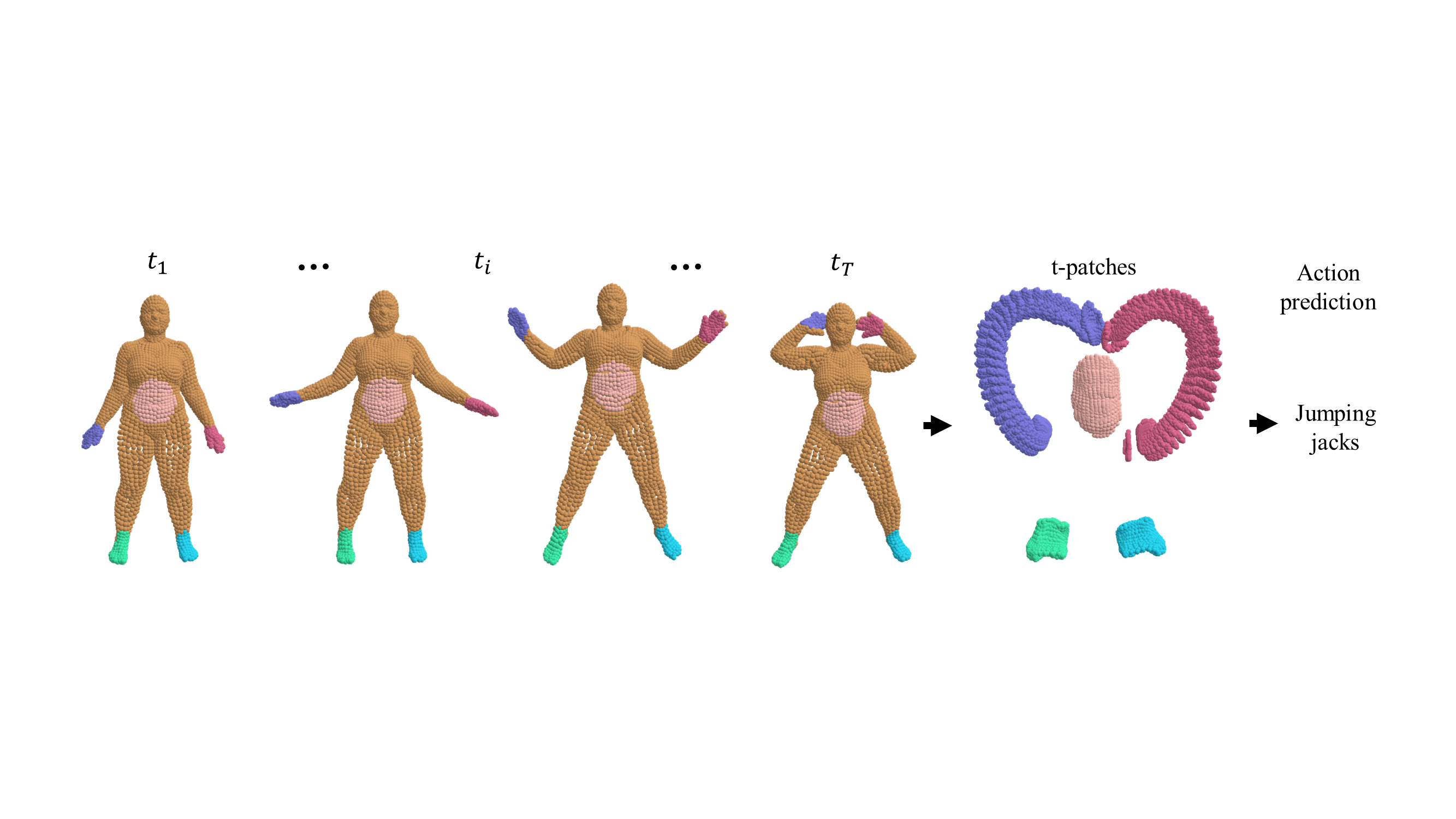}
    \caption{\textbf{t-patches for action recognition.} We propose a new representation for dynamic 3D point clouds. Termed \emph{t-patches}, these are locally evolving point cloud sets aggregated over time. Learning features over t-patches provides an improved temporal point cloud representation for action understanding.}
    \label{fig:teaser}
    \vspace{-0.9cm}
}

\title{3DInAction: Understanding Human Actions in 3D Point Clouds}

\author{Yizhak Ben-Shabat$^{1,2}$ \qquad Oren Shrout$^{2}$ \qquad Stephen Gould$^{1}$\\
\\
${}^{1}$Australian National University \qquad
${}^{2}$Technion, Israel Institute of Technology\\
\tt\small sitzikbs@technion.ac.il, shrout.oren@campus.technion.ac.il, stephen.gould@anu.edu.au\\
\tt\small { \href{https://github.com/sitzikbs/3dincaction}{https://github.com/sitzikbs/3dincaction}}\\
}

\date{}

\maketitle
\begin{abstract}
We propose a novel method for 3D point cloud action recognition. Understanding human actions in RGB videos has been widely studied in recent years, however, its 3D point cloud counterpart remains under-explored despite the clear value that 3D information may bring. This is mostly due to the inherent limitation of the point cloud data modality---lack of structure, permutation invariance, and varying number of points---which makes it difficult to learn a spatio-temporal representation. 
To address this limitation, we propose the 3DinAction pipeline that first estimates patches moving in time (t-patches) as a key building block, alongside a hierarchical architecture that learns an informative spatio-temporal representation. We show that our method achieves improved performance on existing datasets, including DFAUST and IKEA ASM.
Code is publicly available at
\href{https://github.com/sitzikbs/3dincaction}{https://github.com/sitzikbs/3dincaction}.
\end{abstract}


\section{Introduction}
\label{Sec:intro}

In this paper, we address the task of action recognition from 3D point cloud sequences. We propose a novel pipeline wherein points are grouped into temporally evolving patches that capture discriminative action dynamics. Our work is motivated by the massive growth of online media, mobile and surveillance cameras 
that have enabled the computer vision community to develop many data-driven action-recognition methods~\cite{i3d, feichtenhofer2019slowfast, Qiu_2017p3d, tran2015c3d}, most of which rely on RGB video data. Recently, commodity 3D sensors are gaining increased momentum, however, the 3D point cloud modality for action recognition has yet been under-exploited due to the scarcity of 3D action-labeled data.

In many cases, a pure RGB video-based inference may not be enough and incorporating other modalities like geometry is required.  This is especially necessary for safety critical applications such as autonomous systems, where redundancy is crucial, or in scenarios where the video is heavily degraded (\eg due to poor lighting). Some approaches incorporate geometrical information  implicitly, \eg through intermediate pose estimation~\cite{duan2022posec3d}. This often entails  extra steps that require more time and resources and is still limited to video input. Therefore a more explicit approach is desirable.

3D sensors provide an alternative modality in the form of point clouds sampled on the environment. Despite the vast research on 3D vision and learning, even static 3D point cloud datasets are significantly smaller than their RGB image counterparts due to difficulties in collecting and labeling. 3D point cloud sequence databases are even smaller, 

making it more difficult to learn a meaningful 3D action representation.
Furthermore, learning a point cloud representation still remains an active research field because point clouds are unstructured, unordered, and may contain a varying number of points. Learning a temporal point cloud representation is even more challenging since, unlike pixels, there is no one-to-one point correspondence through time.   

We address these challenges and propose the 3DinAction pipeline for 3D point cloud action recognition. 
In our pipeline, we first extract local temporal point patches (t-patches) that reflect a point region's motion in time, see \figref{fig:teaser}. We then learn a t-patch representation using a novel hierarchical architecture that incorporates spatial features in the temporal domain. We finally get an action prediction for each frame in a sequence by aggregating multiple t-patch representations. This pipeline overcomes the need for ground truth point temporal correspondence, grid structure, point order, and a fixed number of points in each frame. Intuitively, patches reflect local surface deformation and are more robust to point correspondence errors. 

 We conduct extended experiments to evaluate the performance of our approach compared to existing SoTA methods and show that 3DinAction provides significant performance gains of $13\%$ and $7\%$ in accuracy on DFAUST and IKEA ASM, respectively.

The key contributions of our work are as follows: 
\begin{itemize}
    \item A novel representation for dynamically evolving local point cloud sets termed t-patches.
    \item A hierarchical architecture that produces an informative spatio-temporal representation for sequences of point clouds. 
\end{itemize}


\section{Related Work}
\label{Sec:related-work}

\noindent\textbf{Learning 3D point cloud representations. } 
Point clouds pose a challenge for neural networks due to their unstructured and point-wise unordered nature. To address these challenges, several approaches have been proposed. PointNet~\cite{qi2017pointnet, qi2017pointnet++} uses permutation-invariant operators, such as pointwise MLPs and pooling layers, to aggregate features across a point set.
Some approaches construct a graph from the point set. DGCNN~\cite{wang2019dynamic} applies message passing and performs graph convolutions on kNN graphs, KCNet~\cite{shen2018mining} uses kernel correlation and graph pooling, and Kd-Networks~\cite{klokov2017escape} apply multiplicative transformations and share the parameters based on the subdivisions imposed by kd-trees. Alternatively, the structure can be imposed using a grid of voxels~\cite{maturana2015voxnet, zhang2021pvt}, or a grid of Gaussians in 3DmFVNet~\cite{ben20183dmfv}. Another alternative avoids the structure by using Transformer's attention mechanism~\cite{lee2019set, Zhao_2021_ICCV} . 
For a comprehensive survey of point cloud architectures please see~\cite{guo2020deep}.

Recently, various factors that can impact the training of different architectures have been investigated~\cite{goyal2021revisiting, qian2022pointnext}. This includes exploring data augmentation strategies and loss functions that are not specific to a particular architecture. The results of this study showed that older PointNet-based architectures ~\cite{qi2017pointnet, qi2017pointnet++} can perform comparably to newer architectures with minor changes.

All of the above methods deal with static, single-frame, or single-shape point clouds. In this work, the input is a temporal point cloud where a representation for a short sequence is required and point correspondence between frames is unknown. Therefore extending existing approaches is not trivial. 

\noindent\textbf{Learning temporal 3D point cloud representations. } Temporal point clouds have not been as extensively studied as their static counterparts, in particular for action recognition. Meteornet~\cite{liu2019meteornet} processes 4D points using a PointNet$^{++}$  architecture where they appended a temporal dimension to the spatial coordinates. PSTNet~\cite{fan2021pstnet, fan2021deep} proposed spatio-temporal convolutions and utilized some of the temporal consistency for action recognition.   Similarly, P4Transformer~\cite{fan2021p4transformer} uses 4D convolutions and a transformer for capturing appearance and motion via self-attention. In a follow-up work PST-Transformer~\cite{fan2022psttransformer} employs a video level of self-attention in search for similar points across entire videos and so encodes spatio-temporal structure. Some works attempt to alleviate the full supervision requirement for 3D action recognition. These include self-supervised features learning \cite{wang2021self} by predicting temporal order from a large unlabeled dataset and fine-tuning on a smaller annotated datasets and unsupervised skeleton colorization \cite{yang2021skeleton}. Additional supervised approaches include
MinkowskiNet ~\cite{choy20194d} that uses a 4D spatio-temporal CNN after converting the point clouds to an occupancy grid, 3DV~\cite{wang20203dv} that encodes 3D motion information from depth videos into a compact voxel set, and Kinet~\cite{zhong2022kinet} that implicitly encoded feature level dynamics in feature space by unrolling the normal solver of ST-surfaces. 

The above methods, perform a single classification per clip. In this paper, we focus on a related, and more chllanging, task that requires a prediction per-frame. We propose to convert the point cloud representation into t-patches and use an MLP based hierarchical architecture to get the spatio-temporal representation.

\noindent\textbf{3D action understanding datasets. } 
One of the major driving forces behind the success of learning-based approaches is the availability of annotated data. For the task of 3D point cloud action recognition, there is currently no designated standard dataset, however, some existing datasets may be extended.
The CAD 60 and CAD 120~\cite{koppula2013learning, sung2012unstructured} datasets include 60 and 120 long-term activity videos of 12 and 10 classes respectively (\eg making cereal, microwave food). These datasets provide raw RGB, skeletons, and depth data however its small scale and long-term focus limit its effectiveness.  
The NTU RGB+D 60~\cite{shahroudy2016ntu} and NTU RGB+D 120 ~\cite{liu2019ntu} provide
$\sim$56K and $\sim$114K clips containing 60 and 120 actions classes respectively, \eg taking off a jacket, taking a selfie. They provide three different simultaneous RGB views,
IR and depth streams as well as 3D skeletons. While these datasets can be considered large-scale, their contrived nature makes recent skeleton-based methods (\eg~\cite{duan2022posec3d}) perform well, making a prior-free approach difficult to justify.
The MSR-Action3D dataset~\cite{li2010msraction3d} includes 20 action classes performed by 10 subjects for a total of 567 depth map sequences, collected using a Kinect v1 device (23K frames). The sequences in this dataset are very short and therefore using it to evaluate learning-based approaches provides a limited indication of generalization. The above datasets provide per clip action annotations. 

 Some datasets inherently provide per-frame annotations. The IKEA ASM dataset~\cite{ikeaasm} provides 371 videos clipped into $~$31K clips. It contains 33 action classes related to furniture assembly, annotated per frame. This dataset provides several modalities including three RGB views, and Depth. It is an extremely challenging dataset since the human assembler is often occluded and presents very unique assembly poses. It is also very imbalanced since different assembly actions have different duration and may repeat multiple times within the same assembly. Although it was designed for video action recognition, its challenges are the core reasons for choosing to extend it to the point cloud action recognition task. 
 The DFAUST dataset~\cite{dfaust:CVPR:2017} provides high-resolution 4D scans of human subjects in motion. It includes 14 action categories with over 100 dynamic scans of 10 subjects (1:1 male-to-female ratio) with varying body shapes represented as registrations of aligned meshes, therefore an extension to our task is straightforward. 
One particularly important feature of this dataset is the GT point correspondences throughout the sequence \ie it is possible to follow each point's movement through time.  While this dataset is not as large-scale as others, it provides ground truth information (correspondence) that most other collected datasets do not. Therefore, we extend this dataset to 3D point cloud action recognition and use it as a testbed for many ablation studies (see \secref{sec:ablation_study}).

\section{3DinAction pipeline}
\label{Sec:approach}
\begin{figure*}[t]
    \centering
    \includegraphics[width=0.98\linewidth,trim={0cm 1.5cm 0cm 0cm},clip]{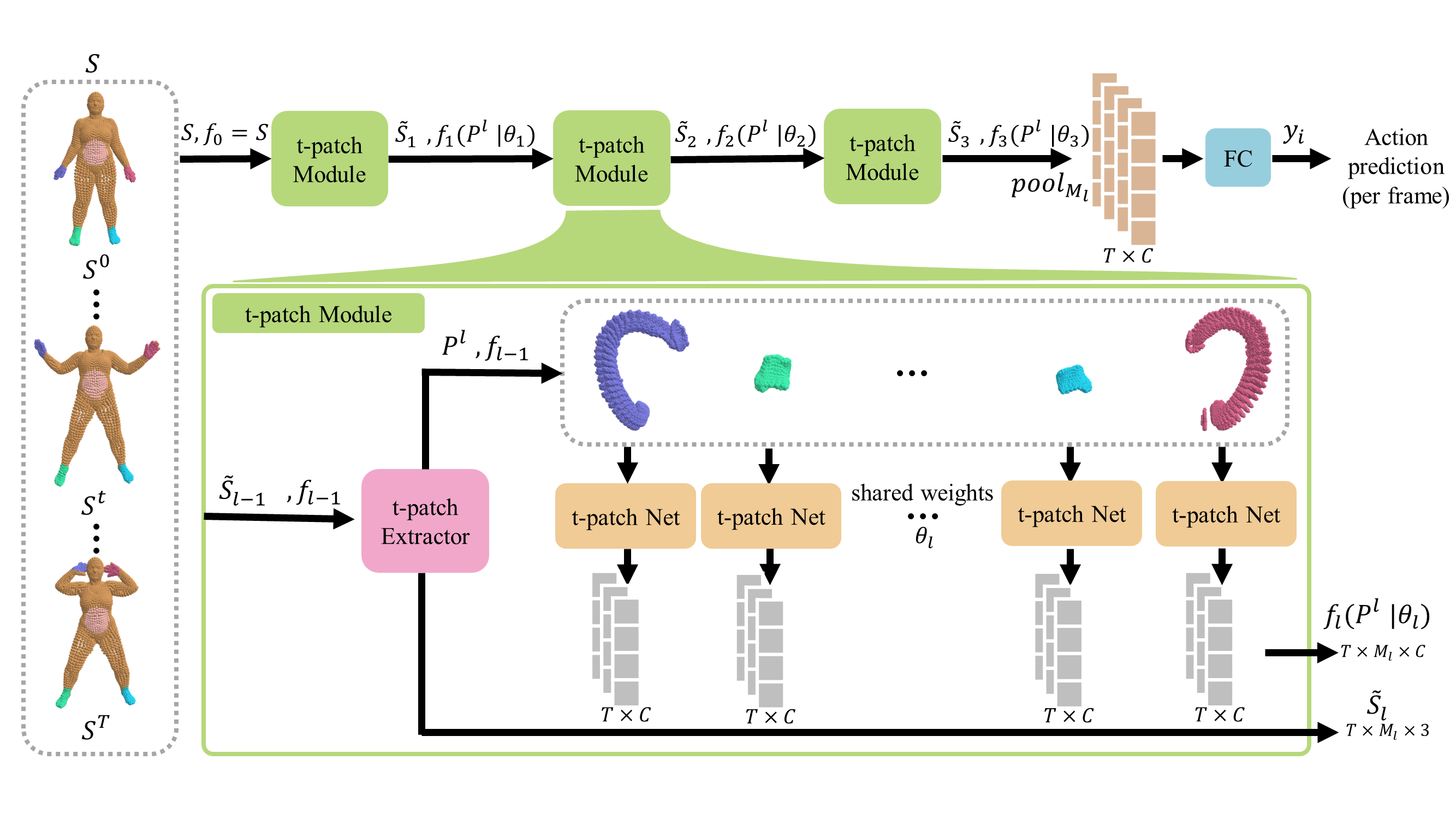}
    \caption{\textbf{3DinAction pipeline.} Given a sequence of point clouds, a set of t-patches is extracted. The t-patches are fed into a neural network to output an embedding vector. This is done hierarchically  until finally the global t-patch vectors are pooled to get a per-frame point cloud embedding which is then fed into a classifier to output an action prediction per frame.}
    \label{fig:pipeline}
\end{figure*}

Our 3DinAction pipeline is illustrated in \figref{fig:pipeline}. Given a temporal sequence of 3D point clouds we first extract a set of t-patches (\secref{subSec:approach}). We then feed the t-patches into a hierarchical neural network (\secref{subsec:network}) to produce a per-frame high dimensional feature vector representation. Finally,  the feature vectors are fed into a classifier to obtain per-frame predictions. The proposed approach is prior-free (no skeleton extraction required) and therefore general and can be used on different action-understanding datasets. 

\subsection{t-patches}
\label{subSec:approach}

Let $S = \{x_j \in \reals^3 \mid j = 1, \ldots, N\}$ denote a 3D point cloud with $N$ points. In the classic (static) setting, a patch $\Psi_q$ is extracted around some query point $x_q$. For example, the patch $\Psi_q$ may be constructed by finding the $k$-nearest neighbors of $x_q$ in $S$.

In our temporal setting we are given a sequence of point clouds ${\cal S} = \{S^0, \ldots, S^T\}$ composed of point cloud frames $S^t = \{x_j^t \mid j = 1, \ldots, N^t\}$. Here the superscript $t$ is used to denote the index of the point cloud in the sequence. Instead of extracting a patch within a single frame, we allow patches to extend temporally, and denote them as \emph{t-patches}.

\begin{definition}
A t-patch $P_q$ is a sequence of point sets indexed by a query point $x_q^0$ and jointly moving in time defined by a pointwise mapping function between patches in consecutive frames. Mathematically, $P_q = \langle \Psi_q^t \rangle_{t=0}^{T}$
where $\Psi_q^0$ is the initial (static) patch and $\Psi_q^t = \Phi(\Psi_q^{t-1})$ is the patch at time $t$ where   $\Phi$ is a pointwise mapping function.
\end{definition}

In practice, it is difficult to find a reliable mapping function $\Phi$. Therefore we propose a simplified formulation that, for a given query point $x_q^0$, first extracts a patch  for the first frame $\Psi_q^0$ and then iteratively extracts corresponding patches for the next frames (iterating over time), by using the closest point in the next frame as the new query point. More formally, we specify $\overrightarrow{\Psi}_q^0 \triangleq \Psi_q^0$, $\overrightarrow{\Psi}_q^t = knn(x_q^{t-1}, S^{t})$ and $x_q^t=nn(x_q^{t-1}, S^{t})$ for $t=1,\ldots, T$. Here $knn$ is the $k$ nearest neighbor and $nn$ is nearest neighbor. Then, the simplified t-patch formulation is given by 
\begin{align}
    \overrightarrow{P}_q &= \langle \overrightarrow{\Psi}_q^t \mid t = 0, \ldots, T \rangle
\label{eq:tpach_simplified}
\end{align}

\begin{figure}
    \centering
    \includegraphics[width=\linewidth,trim={4cm 4cm 5cm 4cm},clip]{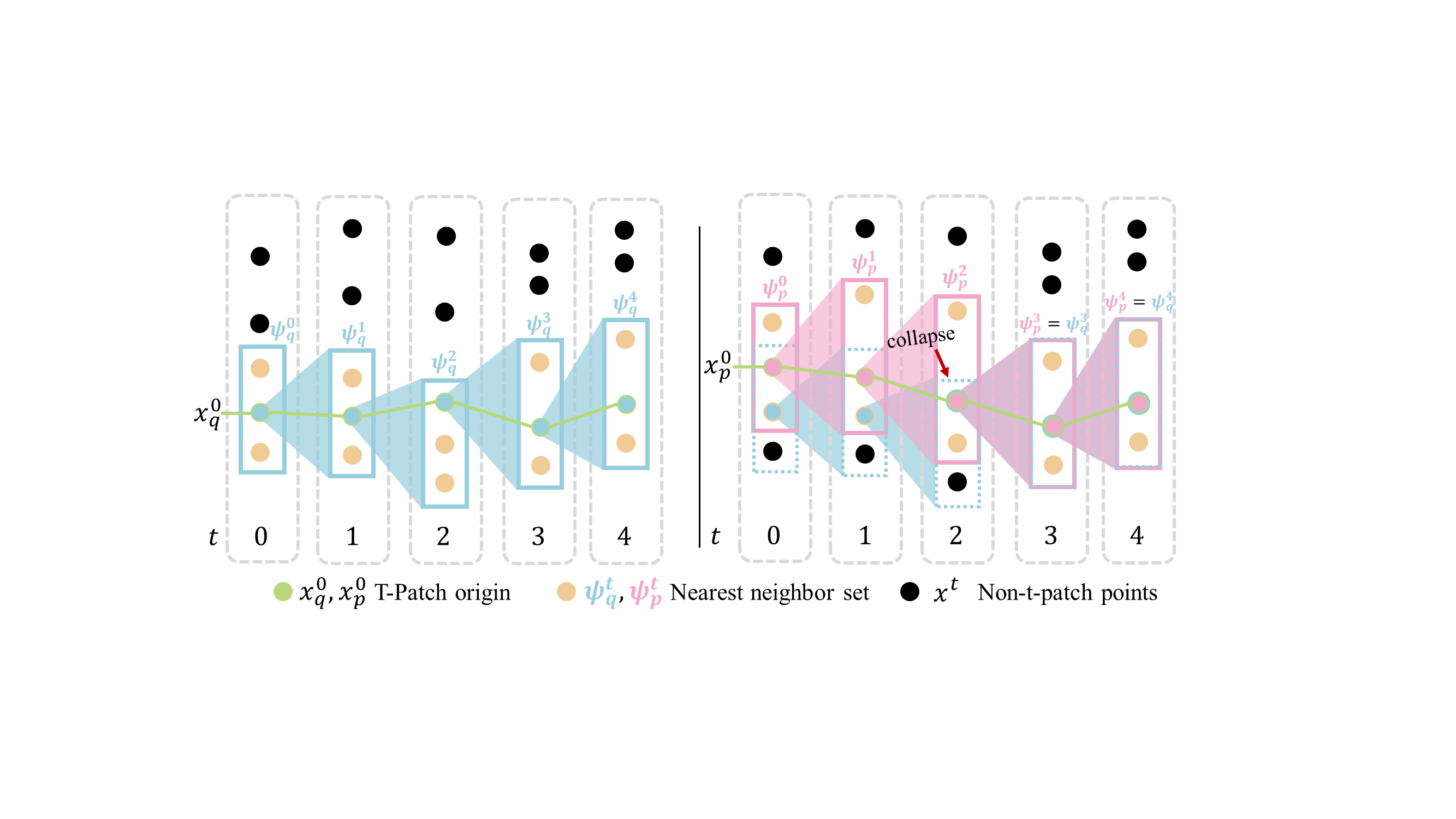}
    \caption{\textbf{t-patch construction and collapse.} Illustration of t-patch construction (left) and collapse (right). Starting from an origin point $x_q^{0}$ we find the nearest neighbours in the next frame iteratively to construct the t-patch subset (non-black points). A collapse happens when two different origin points, $x_q^{0}$ and $x_p^{0}$, have the same nearest neighbour at some time step, $\Psi_p^3=\Psi_q^3$ here.}
    \label{fig:t-patch_extraction}
\end{figure}

See \figref{fig:t-patch_extraction} left for an illustration of the t-patch extraction process. 
Note that if ground truth correspondence is available \textit{knn} can be swapped back to~$\Phi$. However, this does not guarantee improved performance. 

\noindent\textbf{Temporal t-patch collapse.}
The simplified formulation of extracting t-patches inherently suffers from the problem of two or more t-patches collapsing into having the same points after a certain frame. We call this scenario t-patch \emph{temporal collapse}. Temporal collapse can happen whenever $x_{q}^t = x_{p}^t$ for $x_{q}^0 \neq x_{p}^0$. The main issue with temporal collapse is the reduction in point coverage as time progresses, \ie the patches covering the last point cloud have significant overlaps and therefore include fewer points than the first frame and so missing vital data. An illustration of the t-patch collapse problem is available in \figref{fig:t-patch_extraction} (right). To mitigate this issue, we propose two solutions. First, adding small noise to each iteration's query points, \ie ${\overrightarrow\Psi_q^t = knn(x_q^t+\epsilon, S^{t+1})}$  where  ${\epsilon \sim \mathcal{N}(\mu,\sigma^2)}$ is a small Gaussian noise. Second, we propose to construct t-patches from the first to last frame but also in reverse, initializing with $\Psi_q^0$ and $\Psi_q^T$, respectively. We name this variation \textit{bidirectional t-patches}. 
More formally bidirectional t-patches are given by, 

\begin{align}
    \overleftrightarrow{P} &=
    \left( \bigcup_{q} \overrightarrow{P}_q \right) \cup
    \left( \bigcup_{p} \overleftarrow{P}_p \right)
    \label{eq:tpach_bidirectional}
\end{align}
where $\overleftarrow{P}_p$ is defined similarly to $\overrightarrow{P}_q$ but in the reverse direction, i.e., $\overleftarrow{\Psi}_p^T \triangleq \Psi_p^T$ and $\overleftarrow{\Psi}_p^t = knn(x_p^{t+1}, S^t)$ for $t = T-1, \ldots, 0$.
Here,  the final set of t-patches is composed of an equal number of t-patches from both directions. 

\subsection{Hierarchical architecture}
\label{subsec:network}
The proposed architecture is composed of $l$ consecutive t-patch modules. Each module receives a point cloud sequence $S$ as input. The sequence is fed into a t-patch extractor where it undergoes subsampling and t-patch extraction, forming $\tilde S_l$ and $P^{l}$  respectively. Then, the t-patches are fed into t-patch Net, a network that computes a high-dimensional feature vector $f_l$ for each t-patch,   parametrized by $\theta_l$.  The subsampled sequence $\tilde S_l$ and its corresponding t-patch features $f_l$ are then fed into the next t-patch module. These modules form a hierarchy in the sense that each module receives as input a sparser point cloud with a higher dimensional feature vector representing each point (encoding both spatial and temporal information). Note that both the t-patch points and their  features are fed into t-patch Net.

\noindent\textbf{t-patch extractor.}
We first subsample the first frame in the point cloud sequence $S^0$ using farthest point sampling (FPS) to form a set of $M$ query points ${\tilde S^0 = \{x_j^0 \in FPS(S^0, M)\}}$. The set $\tilde S^0$ is used to form the t-patches. Subsampling is required since computing a t-patch for each point is inefficient and unnecessary due to overlaps. After subsampling, we extract $M$ t-patches using \eqnref{eq:tpach_bidirectional} where $q \in \tilde S^0$. The extractor operates on both 3D points and their corresponding features (for modules deeper in the hierarchy). 

\noindent\textbf{Model architecture and t-patch net.}
The t-patch network computes a high dimensional representation for each t-patch. The t-patch Net architecture is composed of several MLP layers operating on the non-temporal dimensions (sharing weights across points) followed by a convolutional layer operating on both the temporal and feature dimensions. Note that the network weights are also shared across t-patches. The output of each t-patch Net is a vector for each frame.  The final frame representation is obtained by aggregating all of the t-patch features using a max pooling operation \ie $maxpool_{M_l}(f_3)$. 
This representation is then fed into a classifier consisting of three fully connected layers with temporal smoothing and softmax to output the final action prediction. 
To train the network we use the same losses of RGB based approaches~\cite{ikeaasm, i3d} which include a per-frame prediction cross entropy loss and a per-sequence prediction cross entropy loss (summed and weighted evenly) $L_{total} = L_{frame} + L_{seq}$. For full details see supplemental.

\section{Experiments}
\label{Sec:results}
We evaluate the performance of our approach on three datasets. The results show that the 3DinAction pipeline outperforms all baselines in DFAUST~\cite{dfaust:CVPR:2017} and IKEA ASM~\cite{ikeaasm} and is comparable in MSR-Action 3D~\cite{li2010msraction3d}. We then conduct an ablation study for selecting parameters and t-patch extraction method showing that adding jitter and bidirectional t-patches is beneficial. Finally, we report time performance and show the tradeoff between performance and inference time. For more results and experiments, see supplemental material. 

\noindent\textbf{Baselines and evaluation metrics.}
For evaluation, we report several standard metrics~\cite{activitynet}: the top1 and top3 frame-wise accuracy are the de facto standard for action classification.  We compute it by summing the number of correctly classified frames and dividing by the total number of frames in each video and then averaging over all videos in the test set. Additionally, since some of the datasets are imbalanced and may contain different actions for each frame in a clip, we also report the macro-recall by separately computing recall for each category and then averaging (macro). Finally, we report the mean average precision (mAP) since all untrimmed videos contain multiple action labels.

For DFAUST and IKEA ASM we report static methods PointNet~\cite{qi2017pointnet},   PointNet$^{++}$~\cite{qi2017pointnet++}, and Set Transformer~\cite{lee2019settransformer} by applying them on each point cloud frame individually. Additionally, we report temporal methods like PSTNet~\cite{fan2021pstnet} and also implemented a temporal smoothing version of each static method (PoinNet+TS, Pointnet$^{++}$+TS, and Set Transformer+TS respectively) by learning the weights of a convolutional layer over the temporal dimension. Temporal smoothing aims to provide a naive baseline for utilizing temporal information in addition to spatial information.
Note that in all experiments, unless otherwise specified, our method uses the simplified formulation with jitter and bidirectional t-patches. 

\subsection{Experiments on DFAUST dataset}
We extend the DFAUST dataset for the task of action recognition and show that the proposed approach outperforms other methods (see \tabref{tab:results:baseline:action_segmentation_dfaust_fps}).

\noindent\textbf{DFAUST dataset~\cite{dfaust:CVPR:2017}.} 
We extended the DFAUST dataset to our task by subdividing it into clips of 64 frames with train and test human subjects. The split was constructed so no subject will appear in both training and test set as well as guarantee that all actions appear in both. The train and test sets contain 76 full-length sequences (395 clips, and $\sim$25K frames) and 53 sequences (313 clips, and $\sim$20K frames) respectively.  Each point cloud frame contains 6890 points. These points are mesh vertices and therefore the density varies greatly (\eg very dense on the face, hands, and feet and sparser on the legs). For all baselines, we  sampled a set of 1024 points using the farthest point sampling algorithm to provide a more uniform set of points. For this dataset, all frames in a clip have the same label. Note that not all actions are performed by all subjects. For the full action list and dataset statistics, see the supplemental. 

\noindent\textbf{Results.} 
The results,  reported in \tabref{tab:results:baseline:action_segmentation_dfaust_fps}, show that our proposed approach outperforms all baselines by a large margin. It also shows that temporal smoothing boosts performance significantly for all static baselines. 
Additionally, to explore the influence of our simplified $ knn$-based temporal point mapping, we used the GT point correspondence to match the consecutive t-patch origin points and report the results as another baseline (Ours + GT corr). The results show that there is a mAP performance gain with GT correspondence, however, it is limited. Note that in most datasets, this GT correspondence is not available. Finally, we also experimented with a Transformer architecture to process the t-patch learned representations and show that it does not provide additional performance boost. This may be attributed to the dataset size.

\begin{table}[] 
    \centering
    \begin{tabular}{l c c c c}
         \toprule
            \multirow{2}{*}{\textbf{Method}} &  \multicolumn{2}{c}{\textbf{Frame acc.}}    \\
            & \textbf{top 1} & \textbf{top 3} &  \textbf{mAP}\\
            \hline
            3DmFVNet~\cite{ben20183dmfv} & 60.86 & 87.68 &  0.7171\\
            PointNet~\cite{qi2017pointnet} & 65.67 & 86.44 & 0.7161 \\
            PointNet$^{++}$~\cite{qi2017pointnet++} & 58.51 & 88.28 & 0.5842 \\

            Set Transformer~\cite{lee2019settransformer} & 52.27 & 81.98 & 0.6209 \\
            \hline
            PoinNet~\cite{qi2017pointnet} + TS & 74.10 & 94.00 & 0.7863 \\
            PointNet$^{++}$~\cite{qi2017pointnet++} + TS & 67.88 & 86.21 & 0.7563\\
            Set Transformer~\cite{lee2019settransformer} + TS & 62.95 & 90.33 & 0.7322 \\
            PSTNet~\cite{fan2021pstnet} & 50.70 & 78.28 &  0.6490\\
            \hline
            Ours + GT corr & \underline{77.67} & \underline{95.38} & \textbf{0.8762} \\
            Ours +  Transformer & 77.09 & 93.7 & 77.49 \\ 
            Ours & \textbf{87.26} & \textbf{99.26} & \underline{0.8616} \\
         \bottomrule
    \end{tabular}
    \caption{\textbf{Action recognition results on DFAUST.} Reporting frame-wise accuracy and mean average precision. Ours outperforms all baselines by a large margin.}
    \label{tab:results:baseline:action_segmentation_dfaust_fps}
\end{table}

\begin{figure}
    \centering

    \begin{subfigure}{.32\linewidth}
    \centering
        \includegraphics[width=.99\linewidth, trim={9.5cm 2cm 9.5cm 4cm},clip]{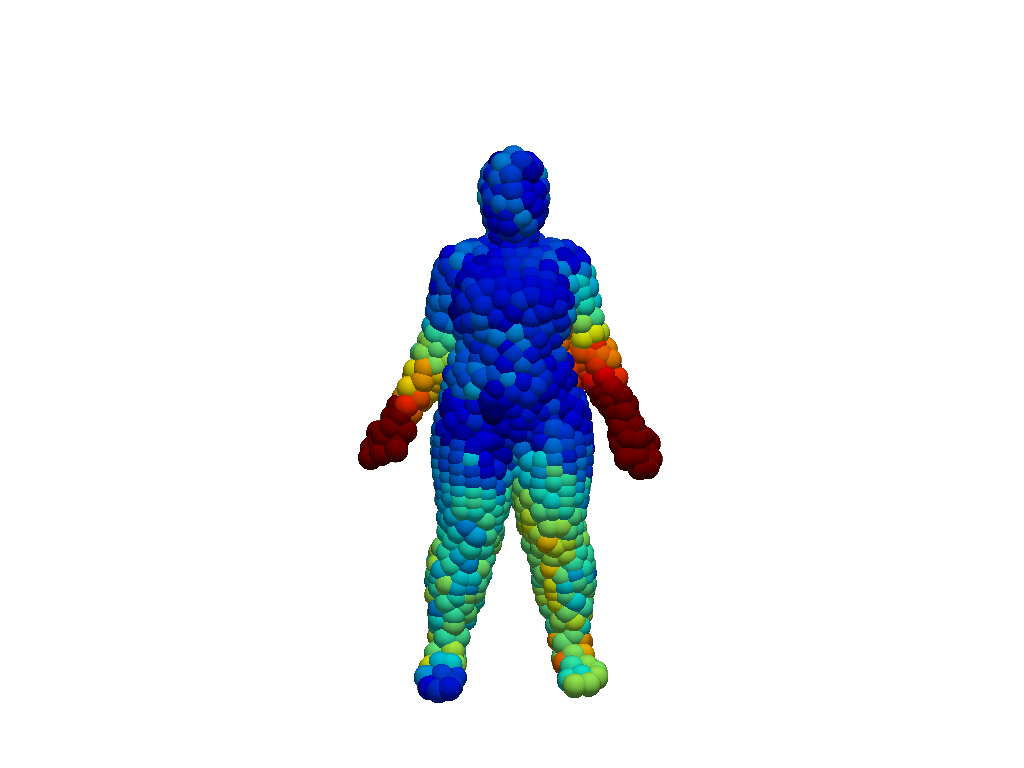}
    \end{subfigure}
    \begin{subfigure}{.32\linewidth}
    \centering
        \includegraphics[width=.99\linewidth, trim={9.5cm 2cm 9.5cm 4cm},clip]{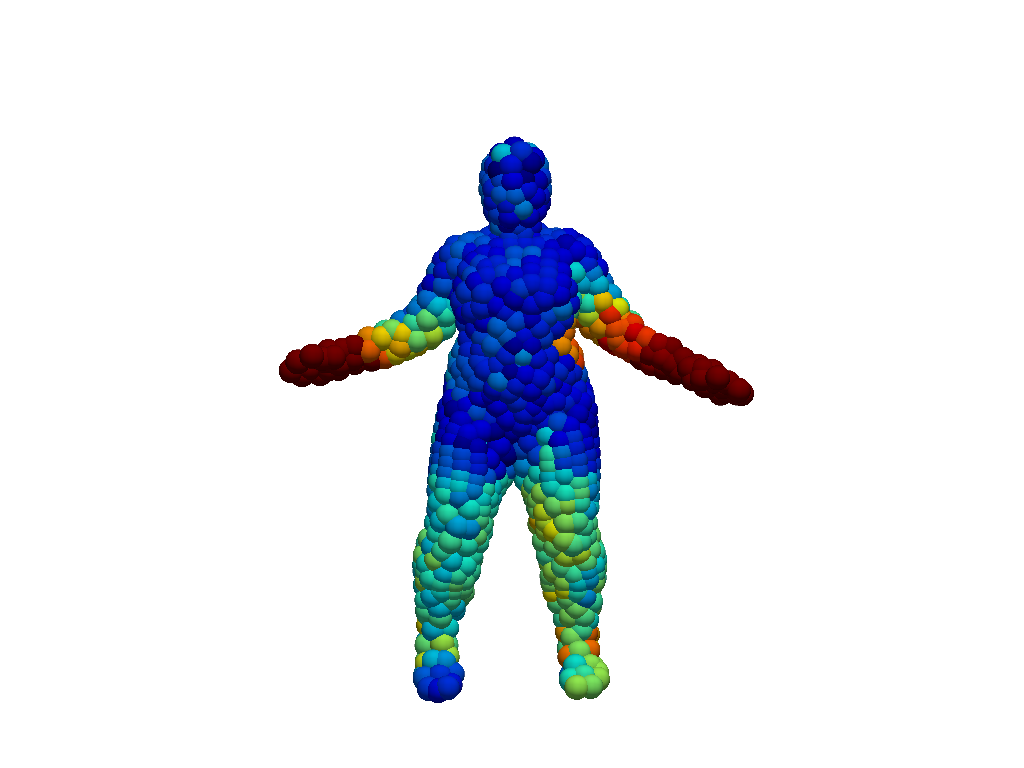}
    \end{subfigure}
        \begin{subfigure}{.32\linewidth}
    \centering
        \includegraphics[width=.99\linewidth, trim={9.5cm 2cm 9.5cm 4cm},clip]{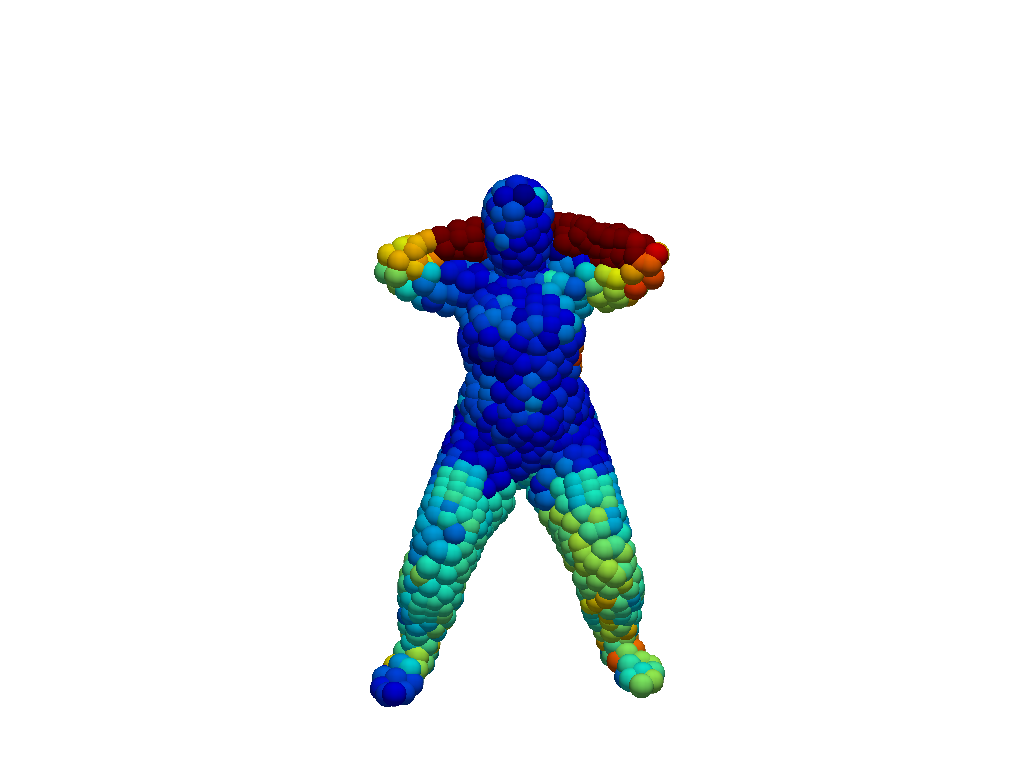}
    \end{subfigure}

            \begin{subfigure}{.32\linewidth}
    \centering
        \includegraphics[width=.99\linewidth, trim={9.5cm 2cm 9.5cm 4cm},clip]{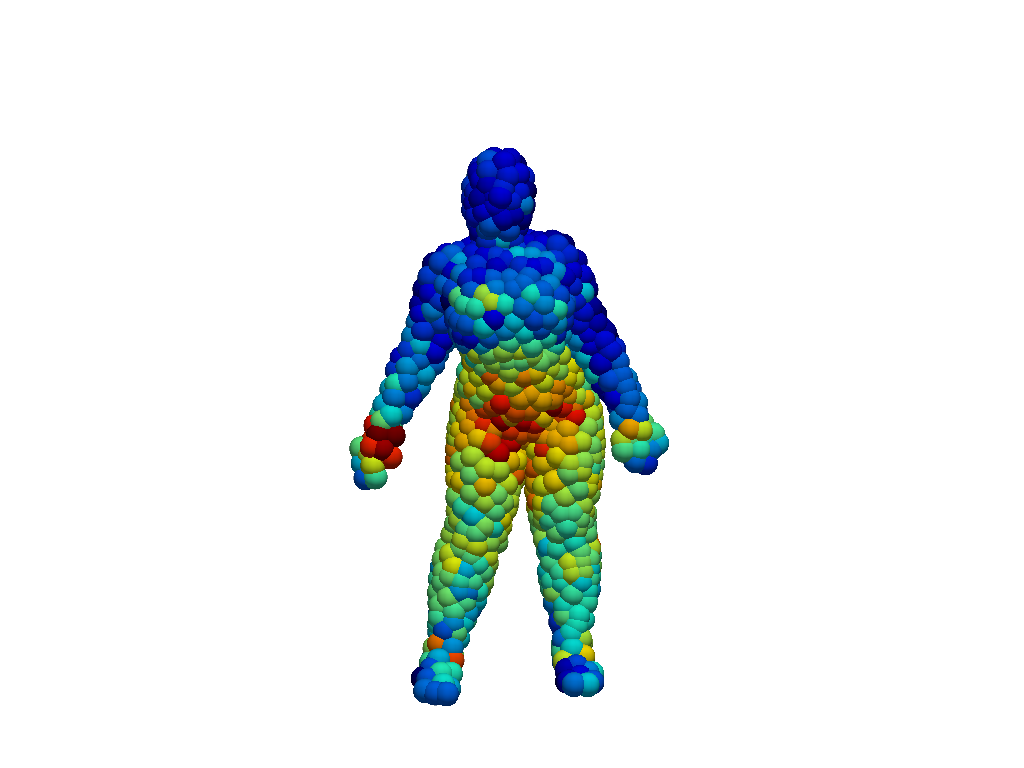}
    \end{subfigure}
    \begin{subfigure}{.32\linewidth}
    \centering
        \includegraphics[width=.99\linewidth, trim={9.5cm 2cm 9.5cm 4cm},clip]{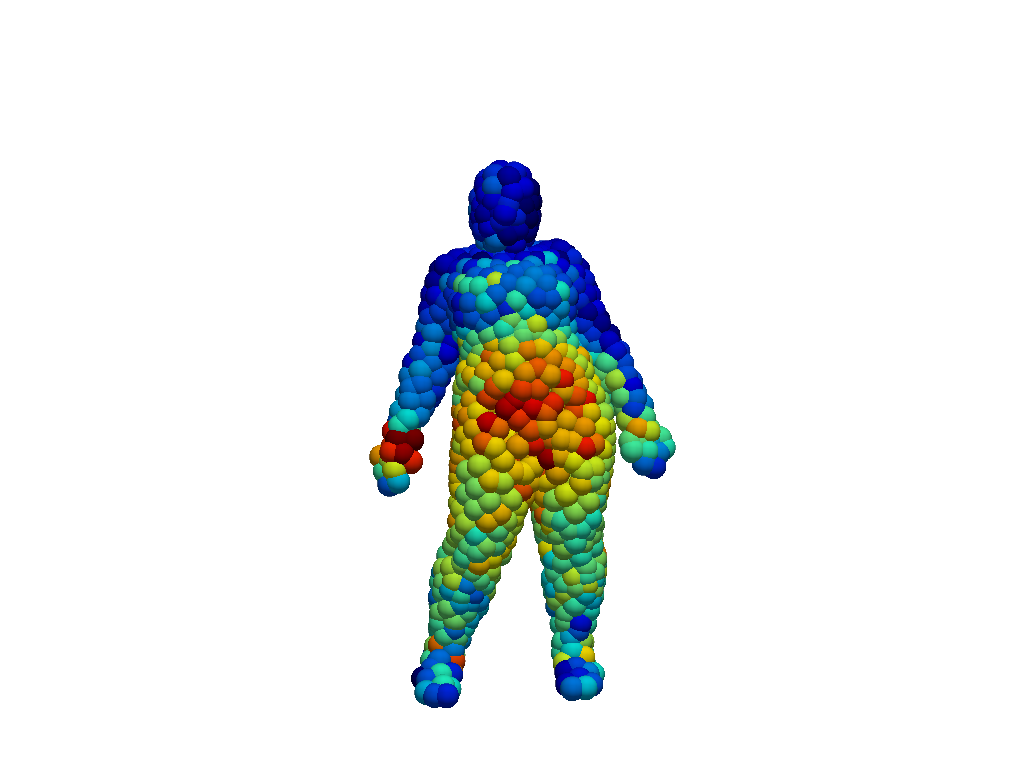}
    \end{subfigure}
        \begin{subfigure}{.32\linewidth}
    \centering
        \includegraphics[width=.99\linewidth, trim={9.5cm 2cm 9.5cm 4cm},clip]{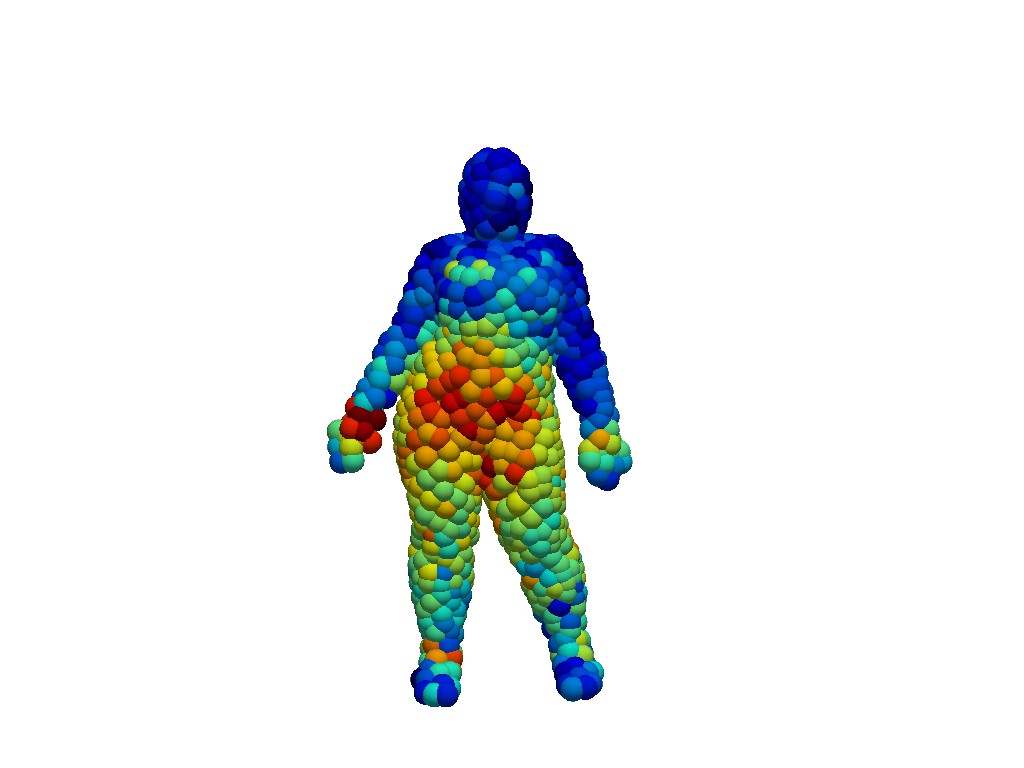}
    \end{subfigure}

        \begin{subfigure}{.32\linewidth}
    \centering
        \includegraphics[width=.99\linewidth, trim={9.5cm 2cm 9.5cm 4cm},clip]{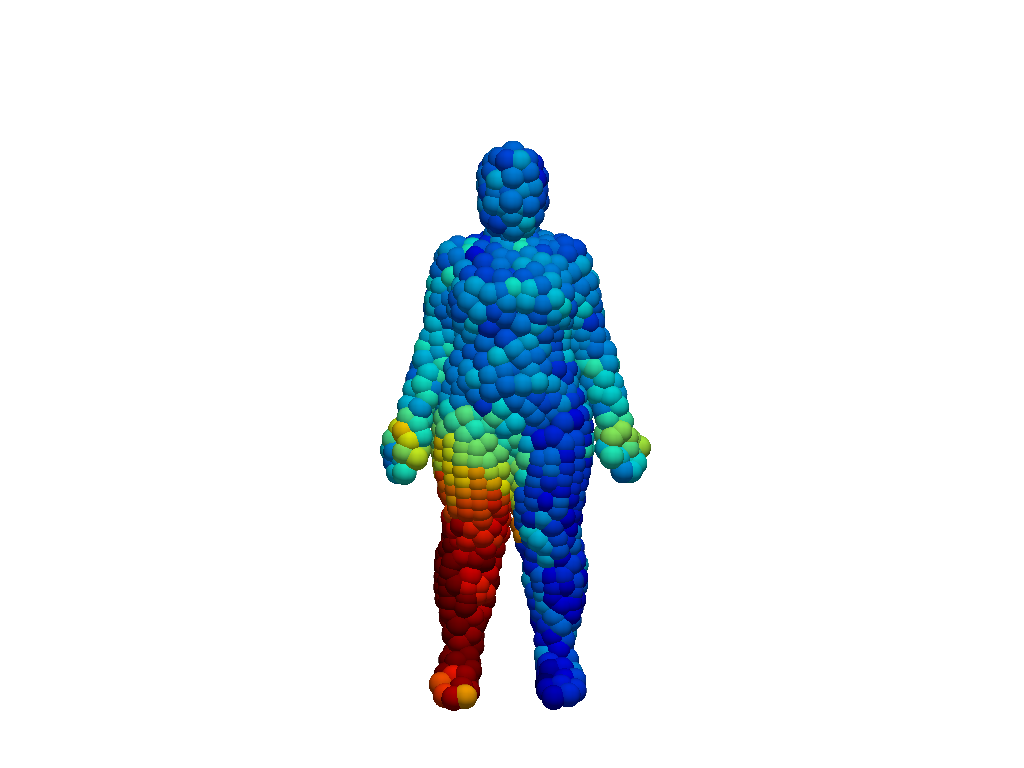}
    \end{subfigure}
    \begin{subfigure}{.32\linewidth}
    \centering
        \includegraphics[width=.99\linewidth, trim={9.5cm 2cm 9.5cm 4cm},clip]{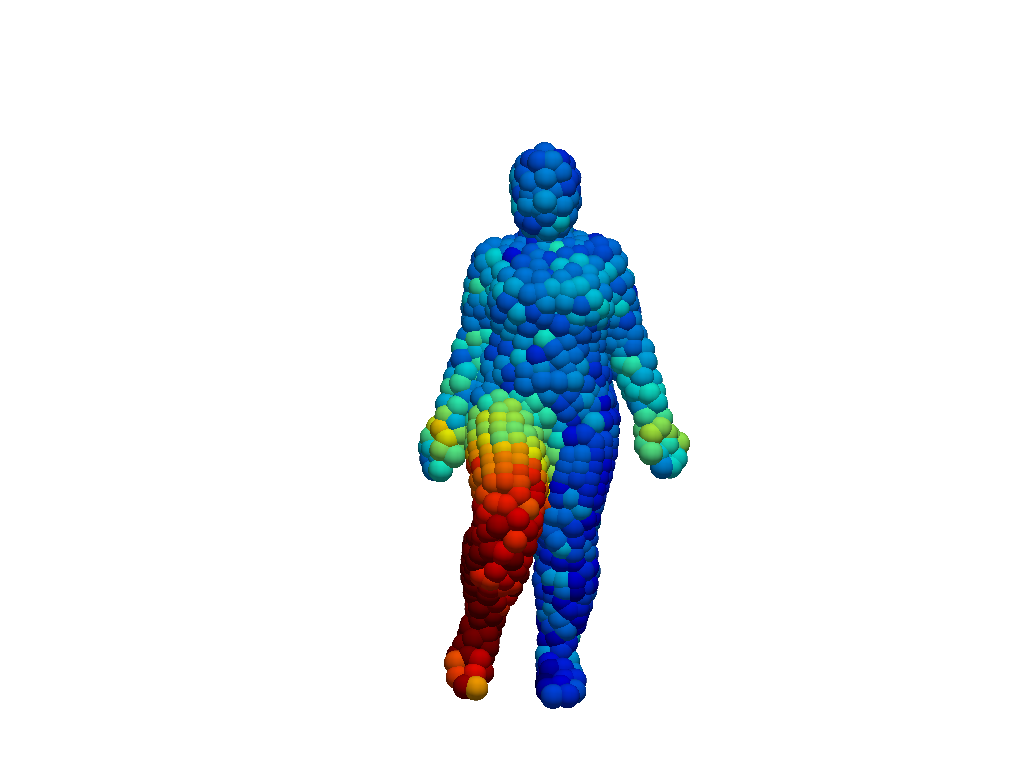}
    \end{subfigure}
        \begin{subfigure}{.32\linewidth}
    \centering
        \includegraphics[width=.99\linewidth, trim={9.5cm 2cm 9.5cm 4cm},clip]{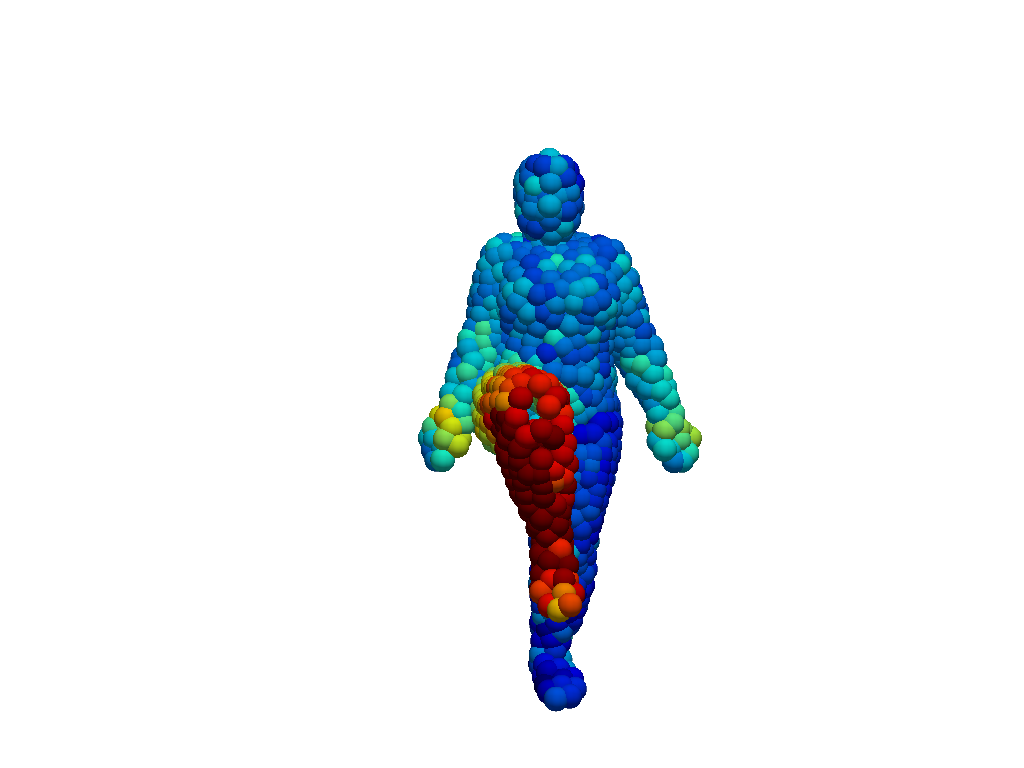}
    \end{subfigure}
    \caption{\textbf{3DinAction GradCAM scores.} The proposed 3DinAction pipeline learns meaningful representations for prominent regions. The presented actions are jumping jacks (top row), hips (middle row), and knees (bottom row). The columns represent progressing time steps from left to right. Colormap indicates high GradCAM scores in red and low scores in blue.}
    \label{fig:gradcam}
\end{figure}

\noindent\textbf{Insight.} We extended the GradCam~\cite{gradcam} approach for our 3DinAction pipeline. Using this approach we get a score per point in each t-patch proportional to its influence on classifying the frame to a given target class. The results in \figref{fig:gradcam} show that, as expected, our approach learns meaningful representations since the most prominent regions are the ones with the informative motion. For example, in the Jumping jacks action (top row) the hands are most prominent as they are making a large and distinct motion. 

\subsection{Experiments on IKEA ASM dataset}
\noindent\textbf{IKEA ASM dataset}~\cite{ikeaasm}. This dataset consists of 371 videos (3M frames) of people assembling IKEA furniture in different indoor environments. It was collected using a Kinect V2 camera and provides camera parameters to reconstruct point clouds in camera coordinates.  It provides action annotation for each frame (33 classes). It is a highly challenging dataset for two main reasons: (1) It is highly imbalanced since some actions have a long duration and occur multiple times in each video (\eg spin leg) and some are shorter and sparser (flip tabletop). (2) The assembly motion includes a lot of self-occlusion as well as subtle movements. The train/test split consists of 254 and 117 full sequences respectively. The split is environment-based (\ie in the test set there is no environment that appeared in the training set). The assembly videos have  an average of $\sim$ 2735 frames per video. The point clouds provided in this dataset are aligned to the camera coordinate frame, posing a challenge for methods that are sensitive to rotations since the camera moves between different scans. 

\noindent\textbf{Results.} 
The results on the IKEA ASM dataset are reported in \tabref{tab:results:baseline:action_segmentation_ikea_asm}. The results show that the proposed 3DinAction pipeline provides a significant performance boost over static approaches and their temporally smooth variants. Additionally, as expected, PointNet and Set Transformer are heavily affected by the variations in coordinate frames. PointNet$^{++}$ on the other hand performs better since it uses local coordinate frames for each local region. All methods show an improved mAP  when using the temporally smooth variant with degradation in frame-wise accuracy due to the dataset imbalance. For this dataset, the top1 metric is not always indicative of the quality of performance because a high top1 is directly correlated with many frames classified as the most common class. Additionally, we compare to pose-based methods reported in \cite{ikeaasm} and show that the proposed approach also outperforms these baselines. See supplementary material for confusion matrices. 

\begin{figure*}
    \centering
    
    \begin{subfigure}{.24\linewidth}
    \centering
        \includegraphics[width=.99\linewidth]{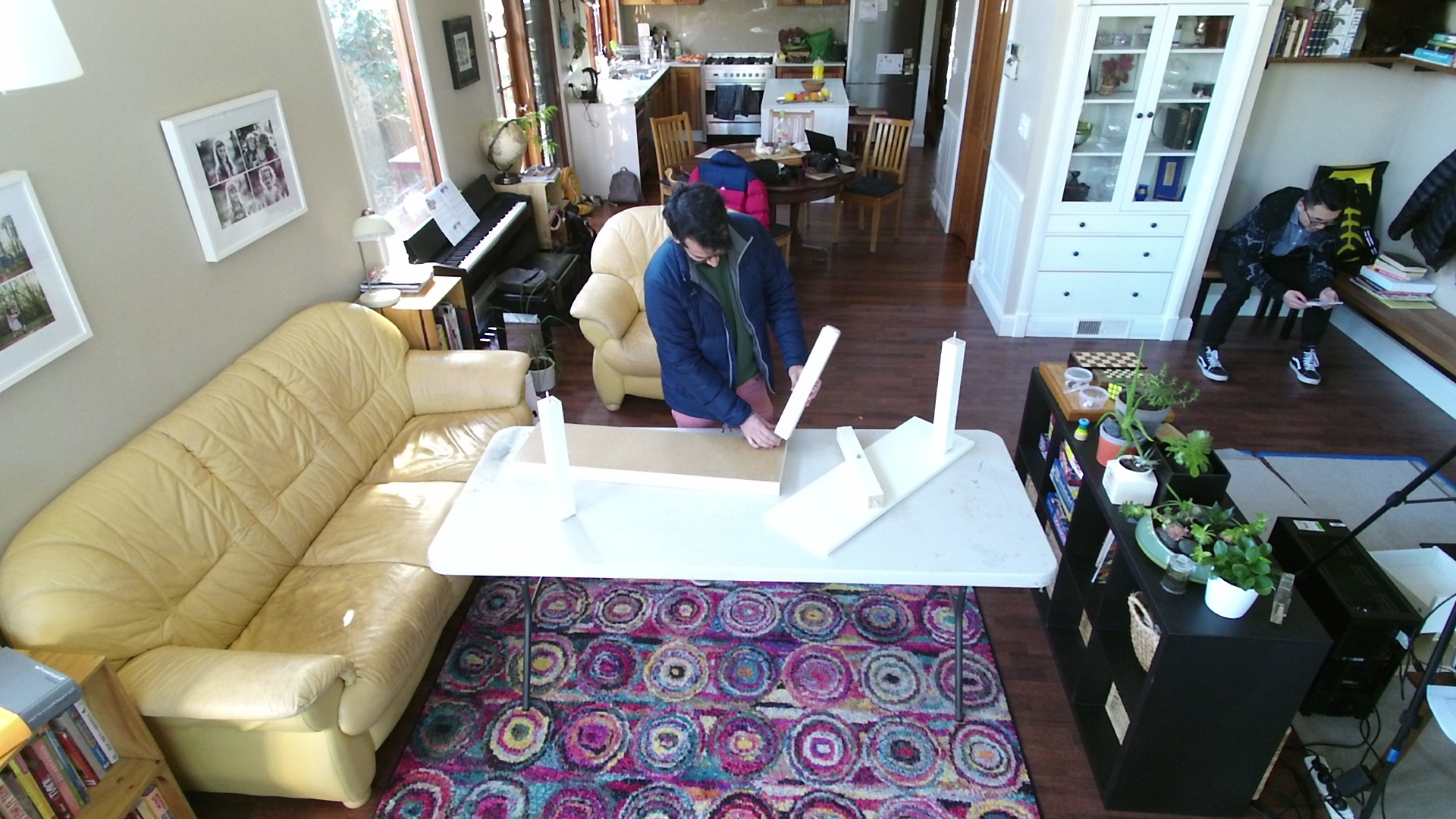}
    \end{subfigure}
    \begin{subfigure}{.24\linewidth}
    \centering
        \includegraphics[width=.99\linewidth]{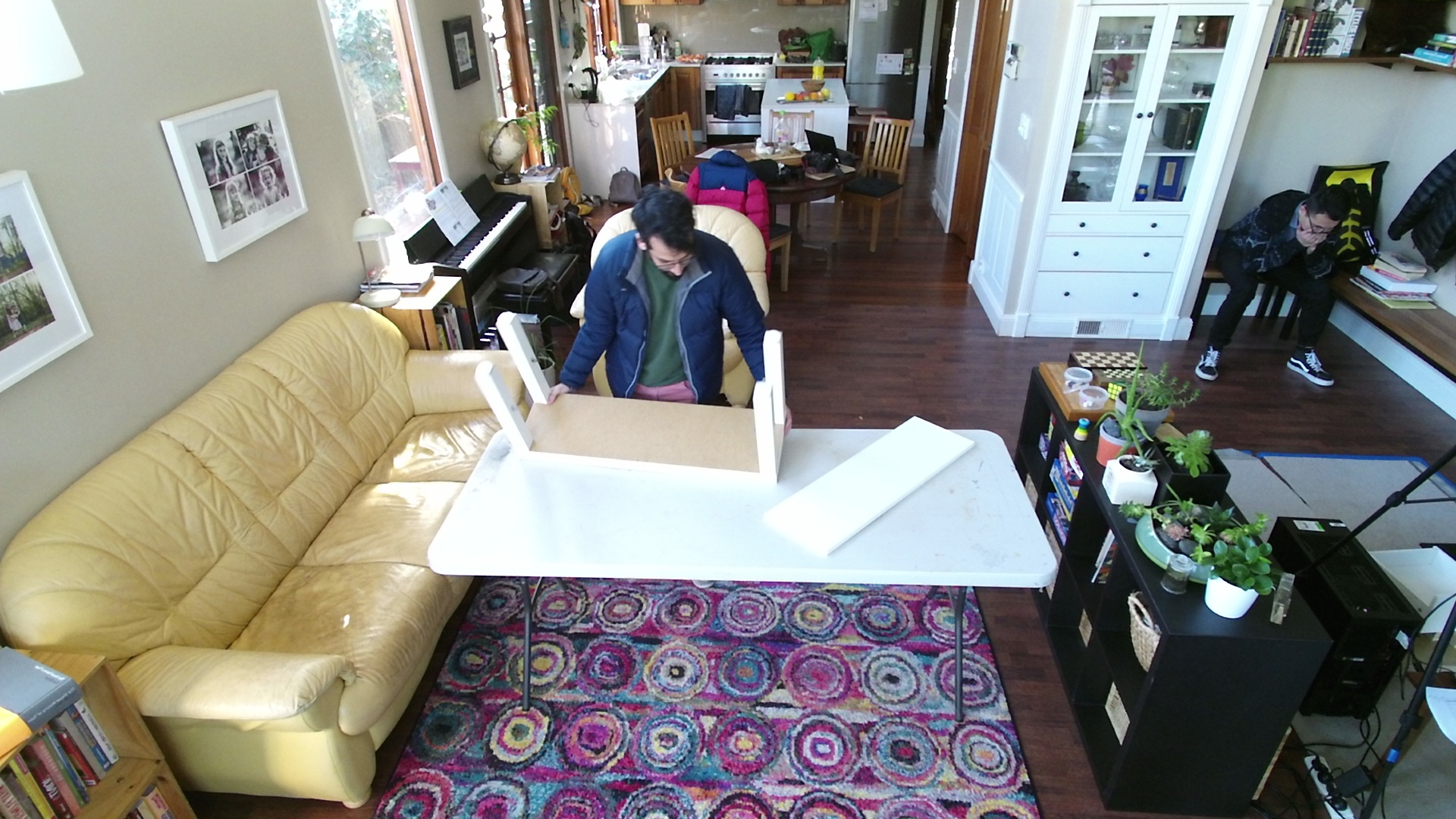}
    \end{subfigure}
        \begin{subfigure}{.24\linewidth}
    \centering
        \includegraphics[width=.99\linewidth]{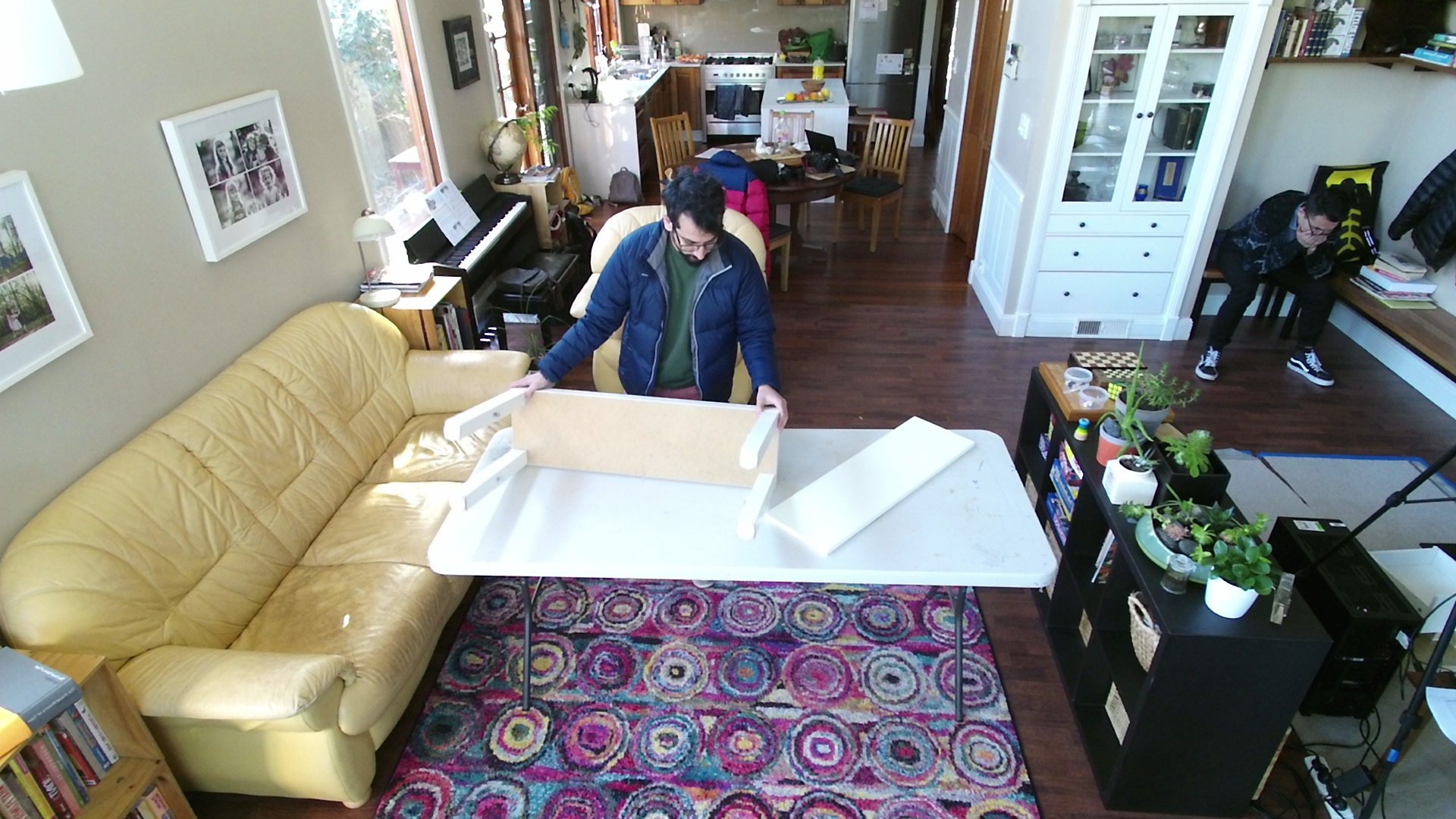}
    \end{subfigure}
    \begin{subfigure}{.24\linewidth}
    \centering
        \includegraphics[width=.99\linewidth]{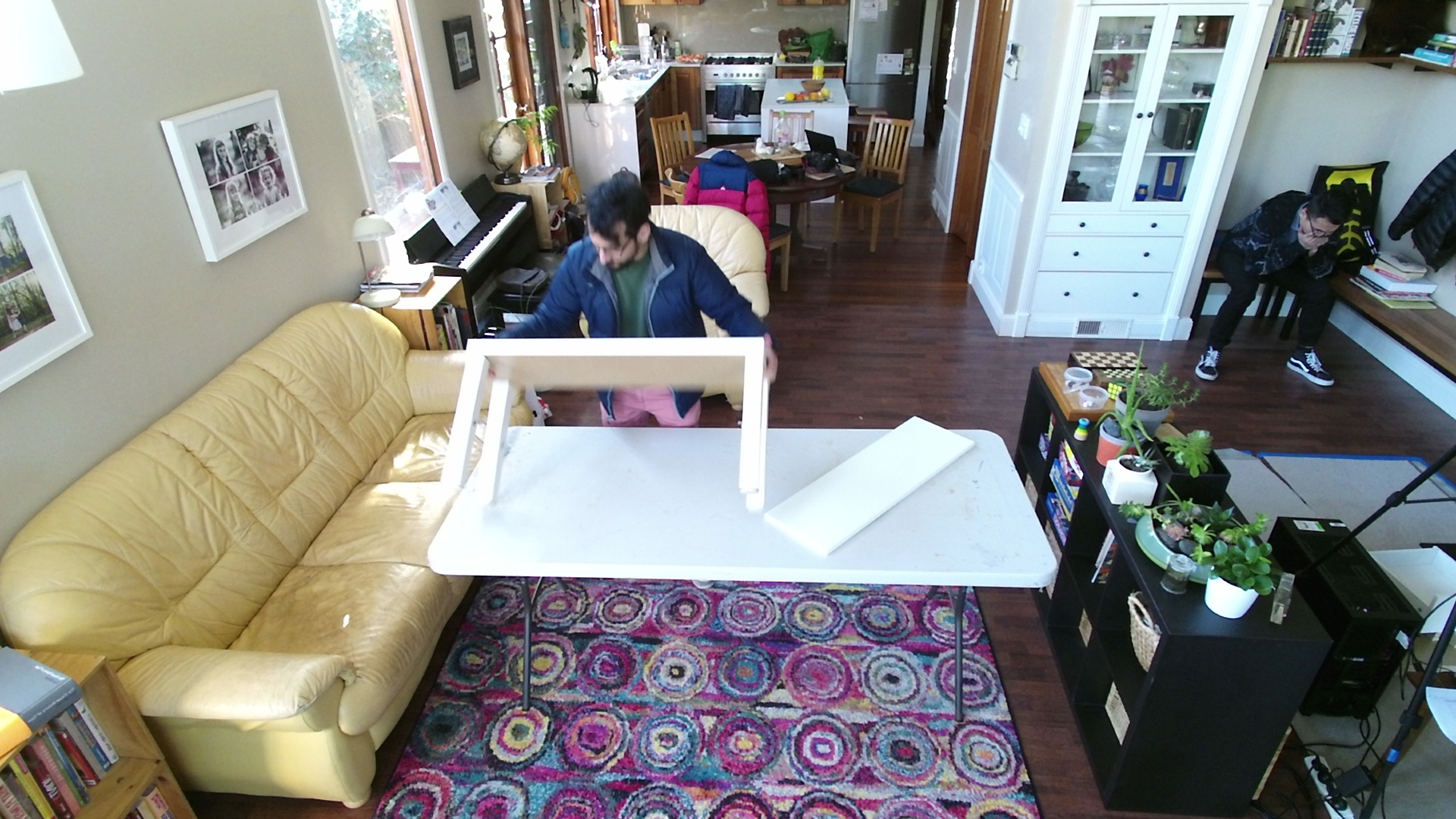}
    \end{subfigure}

        \begin{subfigure}{.24\linewidth}
    \centering
       \includegraphics[width=.99\linewidth, trim={7cm 5cm 3cm 0cm}, clip]{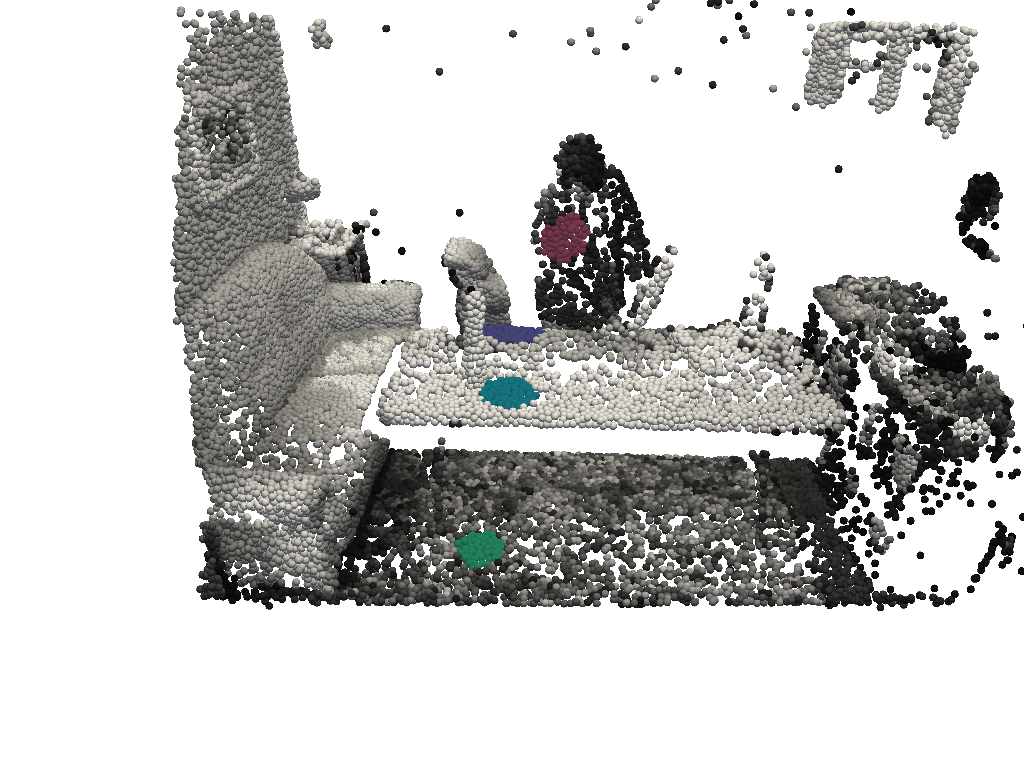}
    \end{subfigure}
    \begin{subfigure}{.24\linewidth}
    \centering
        \includegraphics[width=.99\linewidth, trim={7cm 5cm 3cm 0cm}, clip]{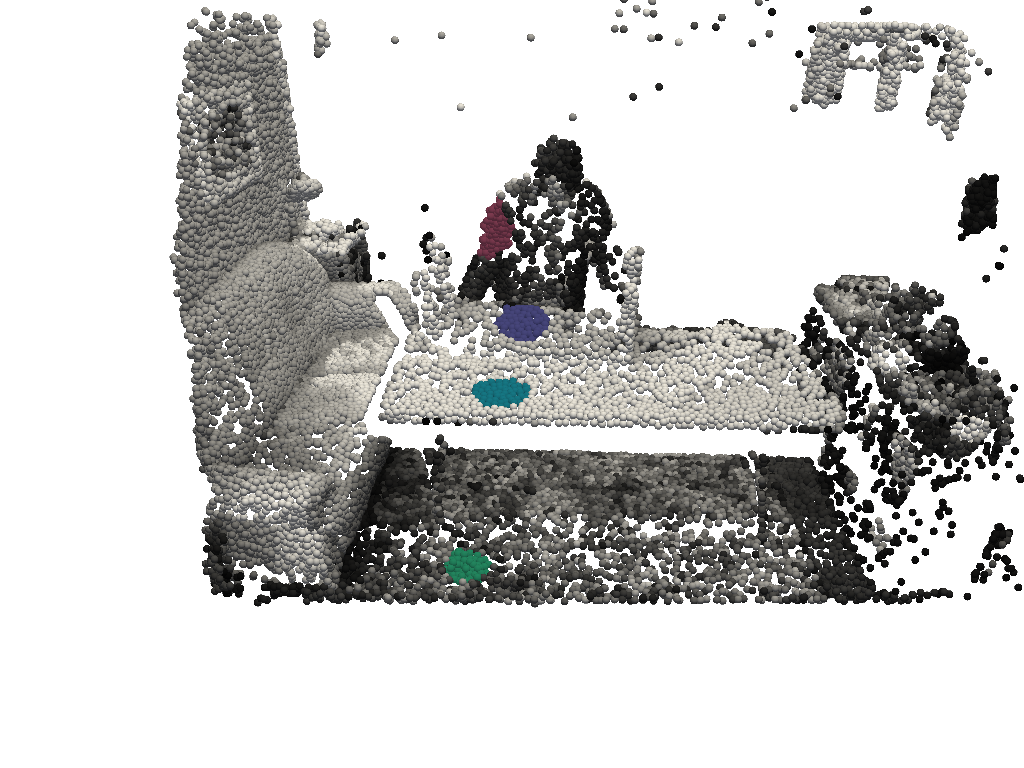}
    \end{subfigure}
        \begin{subfigure}{.24\linewidth}
    \centering
        \includegraphics[width=.99\linewidth, trim={7cm 5cm 3cm 0cm}, clip]{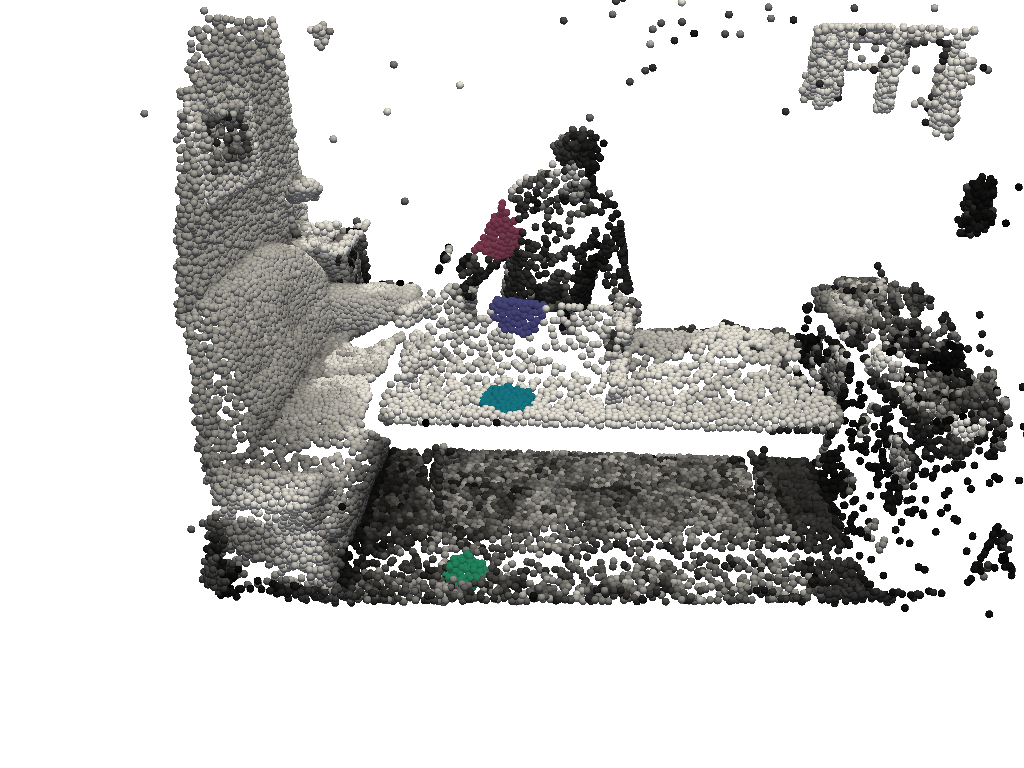}
    \end{subfigure}
    \begin{subfigure}{.24\linewidth}
    \centering
        \includegraphics[width=.99\linewidth, trim={7cm 5cm 3cm 0cm}, clip]{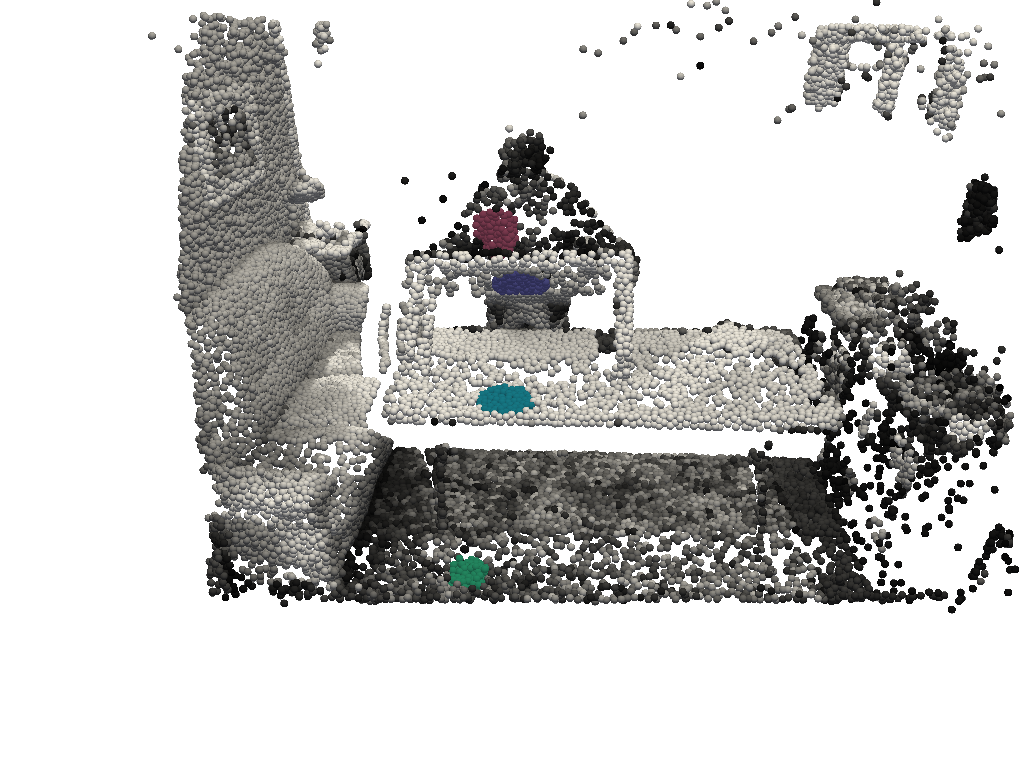}
    \end{subfigure}

    \caption{\textbf{IKEA ASM example with t-patches.}  The flip table action for the TV Bench assembly is visualization including the RGB image (top), and a grayscale 3D point cloud with t-patches (bottom). t-patches are highlighted in color. The \textbf{\textcolor{tpatch_blue}{blue}} is on the moving TV Bench assembly, \textbf{\textcolor{tpatch_pink}{maroon}} is on the moving persons arm,  \textbf{\textcolor{tpatch_cyan}{teal}} is on the static table surface, and \textbf{\textcolor{tpatch_green}{green}} is on the colorful static carpet.}
    \label{fig:IKEA_ASM_flip_table}
\end{figure*}

\noindent\textbf{t-patch intuition and visualization.} 
In \figref{fig:IKEA_ASM_flip_table} we visualize the t-patches for the flip table action in the TV Bench assembly. A set of selected t-patches are highlighted in color demonstrating different types of t-patches and their spatio-temporal changes.  The \textbf{\textcolor{tpatch_blue}{blue}} is on the moving TV Bench assembly, it moves rigidly with the assembly. The \textbf{\textcolor{tpatch_pink}{maroon}} is on the moving person's arm, it exhibits nonrigid motion and deformations through time. The  \textbf{\textcolor{tpatch_cyan}{teal}} is on the static table surface containing some of the TV Bench's points in the first frame but remains static when it moves since its origin query point is on the table. The \textbf{\textcolor{tpatch_green}{green}} is on the static carpet, remaining approximately the same through time. Note that the RGB images are for visualization purposes and are not used in our pipeline.  

\begin{table}[] 
    \centering
    \setlength\tabcolsep{3pt}
    \begin{tabular}{l c c c c}
         \toprule
            \multirow{2}{*}{\textbf{Method}} &  \multicolumn{2}{c}{\textbf{Frame acc.}}    \\
            & \textbf{top 1} & \textbf{top 3} & \textbf{macro} &  \textbf{mAP}\\
            \hline
            PointNet~\cite{qi2017pointnet}  & 4.20 & 19.86 & 5.76 & 0.0346 \\
            PointNet$^{++}$~\cite{qi2017pointnet++}  & 45.97 & 70.10 & 29.48 & 0.1187 \\
            Set Transformer~\cite{lee2019settransformer} & 14.96 & 57.12 & 13.16 & 0.0299 \\
            \hline
            PoinNet~\cite{qi2017pointnet} + TS  & 6.00 & 19.48 & 5.14 & 0.0804 \\
            PointNet$^{++}$~\cite{qi2017pointnet++} +TS & 27.84 & 60.64 & 27.72 & 0.2024 \\
            Set Transformer~\cite{lee2019settransformer} + TS & 9.54 & 36.50 & 10.74 & 0.1471 \\
            PSTNet~\cite{fan2021pstnet} & 17.94 & 52.24 & 17.14 & 0.2016 \\
            \hline
            Human Pose HCN~\cite{ikeaasm}& 39.15 & 65.37 & 28.18 & 0.2232 \\
            Human Pose ST-GCN~\cite{ikeaasm}& 43.4 & 66.29 & 26.54 & 0.1856 \\
            \hline
            Ours  without BD& 45.16 & 72.83 & 35.06 & \textbf{0.2932} \\
             Ours & \textbf{52.91} & \textbf{75.03} & \textbf{38.84} &0.2875 \\ 
         \bottomrule
    \end{tabular}
    \caption{ \textbf{Action classification on IKEA ASM. }The proposed approach provides a significant performance boost over other static and dynamic approaches, including the temporal smoothing (TS). }
    \label{tab:results:baseline:action_segmentation_ikea_asm}
\end{table}

\subsection{Experiments on MSR-Action3D dataset} 
For this dataset, the task is to predict a single class for a sequence of frames (unlike the other datasets where a per-frame prediction is required). To that end, we replace our classifier with a single fully connected layer and max pooled the results over the temporal domain (similar to~\cite{fan2021pstnet}). The results, reported in \tabref{tab:MSRAction3D}, show that all SoTA methods, including the proposed approach, exhibit very similar performance. This is mainly attributed to the small scale of the dataset and the lack of diversity in the action classes. Furthermore, we witnessed that the main performance gap is for frames and sequences where the action is indistinguishable (\eg first few frames of a sequence where no distinguishable action commenced).

\begin{table}[] 
    \centering
    \setlength\tabcolsep{4pt}
    \begin{tabular}{l | c c c c c }
        \toprule
          & \multicolumn{5}{c}{\textbf{\# frames}}   \\
         \textbf{Method} & 4 & 8 & 12 & 16 & 24   \\
         \hline
         PSTNet~\cite{fan2021pstnet} & 81.14 & 83.50 & 87.88 & 89.90 & 91.20  \\
         P4Transformer~\cite{fan2021p4transformer} & 80.13 & 83.17 & 87.54 & 89.56 &  90.94 \\
         PST-Transformer~\cite{fan2022psttransformer} & 81.14 & 83.97 & 88.15 & 91.98 & 93.73  \\
         Kinet~\cite{zhong2022kinet}& 79.80  & 83.84 & 88.53 & 91.92 & 93.27  \\
         \hline
         Ours & 80.47 & 86.20 & 88.22 & 90.57 & 92.23 \\
    \bottomrule
    \end{tabular}
    \caption{\textbf{MSR-Action3D classification results.} Reporting classification accuracy for clips of different lengths. Results show that all methods are comparable since this dataset's scale is limited. }
    \label{tab:MSRAction3D}
\end{table}

\subsection{Ablation study}
\label{sec:ablation_study}
\noindent\textbf{t-patch extraction.} 
We studied the t-patch extraction method and its effects on action recognition on  a noisy version of the DFAUST dataset. The results reported in \tabref{tab:results:baseline:action_segmentation_dfaust_fps_noisy}, show the significance of the t-patch collapse problem and the effectiveness of adding small jitter and bidirectional t-patches to overcome it. In the DFAUST dataset, finding the nearest neighbor between frames provides a ${\sim}96\%$ correspondence accuracy (small motion between frames). Therefore, in this experiment, we augment the dataset once by adding small Gaussian noise to each point in the dataset ($\sigma=0.01$), decreasing the correspondence accuracy to ${\sim}62.4\%$ and introducing multiple t-patch collapse instances as well as increasing the classification difficulty. 

Several variants of the t-patch extraction were explored. The first variation (GT) incorporates the ground truth correspondence into the t-patch extraction. Using this method, there is no t-patch collapse since there is a one-to-one mapping between frames. We expected this to produce an upper bound on the performance, however, surprisingly the results show that this variation is actually inferior to the proposed t-patch approach. We attribute this to the proposed t-patch extraction inherent augmentation caused by the downsampling and nearest neighbor point jitter.
We then continue to explore the proposed approaches for dealing with t-patch collapse which include jitter, \ie adding small noise to each point before finding its nearest neighbor in the next frame, and the bidirectional t-patches that extract patches both from the first to the last frame and from the last to the first frame. The results show that adding jitter is always beneficial and provides a boost in performance. The bidirectional t-patches improve accuracy performance significantly when the data is clean and are comparable when the data is noisy. Note that in both dataset variations, the degradation due to temporal t-patch collapse is low compared to Kinect-based scan data, therefore the bidirectional benefits are not fully utilized. 

\begin{table}[] 
    \centering
         \begin{tabular}{c c c c c c c}
         \toprule

             &  &  &  &\multicolumn{2}{c}{\textbf{Frame acc.}} &    \\
            \textbf{Data} & \textbf{GT} & \textbf{Jitter} & \textbf{BD} & \textbf{top 1} & \textbf{top 3} &  \textbf{mAP}\\
            \hline
             \multirow{4}{*}{clean} & \cmark & \xmark& \xmark & 77.67 & 95.38 & 0.8762 \\  
             & \xmark & \xmark& \xmark & 74.73 & 92.14 & 0.8097\\ 
             & \xmark & \cmark& \xmark & 80.49 & 96.61 & \textbf{0.9023} \\ 
            &  \xmark & \cmark& \cmark & \textbf{87.26} & \textbf{99.26} & 0.8616 \\ 
            \hline
             \multirow{4}{*}{noisy} & \cmark & \xmark& \xmark & 76.08 & 95.50 & \underline{0.9013} \\ 
             & \xmark & \xmark& \xmark & 66.74 & 93.76 & 0.7626 \\ 
            & \xmark & \cmark& \xmark & \textbf{81.83} & \textbf{98.97 }& \textbf{0.9220} \\ 
             &  \xmark & \cmark& \cmark & \underline{80.03} & \underline{97.57} & 0.8975 \\ 

         \bottomrule
    \end{tabular}
    \caption{\textbf{t-patch collapse ablation} on DFAUST. Exploring adding (1) GT - ground truth correspondences, (2) jitter - small Gaussian noise in t-patch construction, and (3) BD - bidirectional t-patches.}
    \label{tab:results:baseline:action_segmentation_dfaust_fps_noisy}
\end{table}

\noindent\textbf{t-patch parameters.} The core parameters for t-patch extraction are the number of neighbors to extract ($k$) and the number of points to subsample ($n$). Here there is a trade-off between complexity and performance \ie when  $k$ and $n$ are small, the input to the model is small accordingly but the overall coverage is reduced and therefore performance is lower. We explored their influence on the noisy DFAUST dataset and report the results in \tabref{tab:ablation:k_n}. The results show that the method is fairly robust to the selection of these parameters, producing comparable results for all. The best performance was obtained for $n=512, k=16$. Surprisingly, the performance slightly degrades when increasing $k$ and $n$ beyond these values. 
This is likely due to the increase in model size, which easily overfits on a dataset of this size.

\begin{table}[] 
    \centering
    \begin{tabular}{c c c c c}
         \toprule
            &&  \multicolumn{2}{c}{\textbf{Frame acc.}}    \\
            \textbf{n} & \textbf{k} & \textbf{top 1} & \textbf{top 3} &  \textbf{mAP} \\
            \hline
            256 & 16 & 76.96 & 97.54 & 0.8430 \\
            512 & 16 & \textbf{80.03} & \underline{97.57} & \textbf{0.8975}\\
            1024 & 16 & 77.30 & \textbf{97.88} &  \underline{0.8507}\\
            512 & 8 & 76.87 & 96.21  & 0.7557 \\
            512 & 32 & \underline{77.91} & 96.60 & 0.7453\\
         \bottomrule
    \end{tabular}
    \caption{\textbf{t-patch parameters ablation}. Results for the number of neighboring points in a patch $k$ and number of downsampled points $n$ show that the method is robust.}
    \label{tab:ablation:k_n}
\end{table}

\noindent\textbf{Time and parameters.}
We report the time performance and the number of parameters of several baselines in \tabref{tab:time_memory_sota}. The results show the tradeoff between performance and time, \ie the temporal approaches exhibit longer processing times and more parameters while performing better.
For the proposed approach, we break down the timing of individual components, namely the t-patch extraction, feature computation, and classifier. The results show that the proposed approach is comparable to PSTNet in time while having more parameters. Interestingly, most of the time is used for extracting the t-patches and not for feature extraction or classification. This is attributed to the farthest point sampling and the sequential \textit{knn} search, both of which could be further optimized for speed. Note that results are average of 50 runs, each with a batch of 4 and 1024 points per frame. 

\begin{table}[] 
    \centering
    \begin{tabular}{l c c }
    \toprule
         \textbf{Method} & \textbf{Time [ms]} & \textbf{\# parameters} \\
         \hline
         PointNet \cite{qi2017pointnet}& 64.49 & 3.5M  \\ 
         PointNet$^{++}$ \cite{qi2017pointnet++}& 23.35  & 1.5M  \\ 
         PSTNet \cite{fan2021pstnet}& 185.92  & 8.3M  \\ 
         \hline
         Ours t-patch extraction & 180.65  & 0 \\ 
         Ours feature computation & 12.50  & 9.8M  \\ 
         Ours classifier & 0.36  & 1.1M  \\ 
         \hline
         Ours &  193.51 & 10.9M  \\ 
         \bottomrule
    \end{tabular}
    \caption{\textbf{Time and parameters.} Temporal methods have more parameters and take longer. 3DinAction time is mostly used to extract t-patches.}
    \label{tab:time_memory_sota}
\end{table}

\noindent\textbf{Limitations.} Since the simplified formulation of t-patch construction uses \textit{knn}, it is sensitive to variations in point densities. A t-patch in a sparse region will occupy a larger volume than a t-patch in a dense region. We use FPS to mitigate this, however, other approaches can be used \eg using neighbors in a fixed radius. Another limitation is data with a very low frame rate or very fast motion since this breaks the assumption that points in consecutive frames are close to each other, and will cause inconsistent t-patch motion. 


\section{Conclusions}
\label{Sec:conclusions}
We introduced the 3DinAction pipeline, a novel method for 3D point cloud action recognition. It showed that the creation of temporal patches is beneficial for finding informative spatio-temporal point representations. 3DinAction has demonstrated a performance boost over SoTA methods. 

This work opens many interesting future directions of research. These include trying to learn the t-patch construction instead of the \textit{knn} selection, imposing stronger temporal structure based on preexisting knowledge and bias (\eg sceneflow or tracking), and exploring using multimodal inputs with this representation (\eg RGB or text). 

\medskip
\noindent
{\small%
{\bf Acknowledgement.}
This project has received funding from the European Union's Horizon 2020 research and innovation programme under the Marie Sklodowska-Curie grant agreement No 893465. We also thank the Microsoft for Azure Credits and NVIDIA Academic Hardware Grant Program for providing high-speed A5000 GPU.}

\bibliographystyle{plain}       
\bibliography{main}
\vfill\clearpage
\pagebreak

\appendix

\setcounter{page}{1}

\twocolumn[
\centering
\Large
\textbf{3DInAction: Understanding human actions from 3D point clouds} \\
\vspace{0.5em}Supplementary Material \\
\vspace{1.0em}
] 
\appendix
\section{Bidirectional t-patches}
\begin{figure}
    \centering
    \includegraphics[width=\linewidth, trim={8cm, 2cm, 8cm, 2cm}, clip]{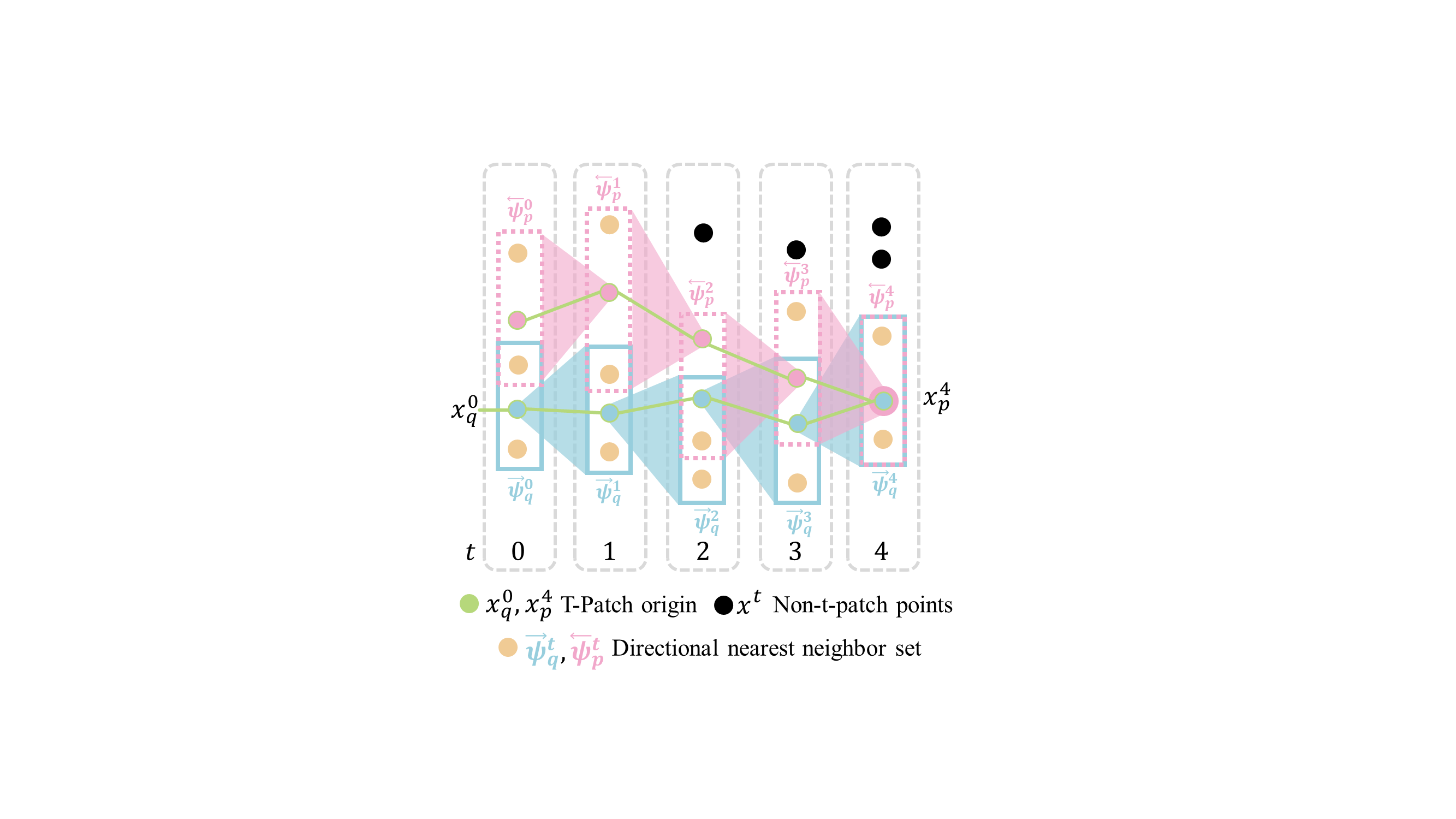}
    \caption{\textbf{Bidirectional t-patch illustration.} t-patches formed from start to finish are presented in light blue  and the reverse-t-patches in pink. Note that the nearest neighbour in one direction is not necessarily the nearest neighbour in reverse (time step $t=4$). }
    \label{fig:bdirectional_tpatch_illustration}
\end{figure}

In \secref{Sec:approach} we presented the t-patches, their construction and the t-patch collapse problem. To mitigate the collapse issue, we proposed a bidirectional t-patch formulation, given in Eq.\eqref{eq:tpach_bidirectional}. In \figref{fig:bdirectional_tpatch_illustration} we present an illustration of this process. The illustration depicts t-patches formed from start to finish in blue $\overrightarrow{\Psi}_q^t$ and in reverse in pink $\overleftarrow{\Psi}_p^t$. Note that the nearest neighbour in one direction is not necessarily the nearest neighbour in reverse, as can be seen in time step $t=4$. The bidirectional t-patches are essential for long sequences in order to keep the coverage ratio and prevent t-patch collapse. 

\section{Pipeline architecture details}
The t-Patch network computes a high-dimensional representation for each t-Patch. The architecture is composed of several MLP layers operating on the non-temporal dimensions (sharing weights across points) followed by a convolutional layer operating on both the temporal and feature dimensions. The network weights are shared across t-patches. 

For the first t-Patch module, 512 t-Patches are extracted and fed into 3 MLPs with dimensions of $(64, 64, 128)$ followed by a 2D temporal convolution with a kernel of $(8, 128)$, \ie operating over all feature channels and weighted averaging 8 consecutive frames. For the second module, 128 points (t-patch centers from the previous extractor) are sampled, the MLP is size is $(128, 128, 256)$ and the temporal kernel is $4$. For the third module, no downsampling is performed, MLPs sizes are $(256, 512, 1024)$. All layers use ReLU activation and batch norm. The final classifier uses 3 fully connected layers of sizes $(512, 256, \#classes)$ with a drop out after the first and second layers with a drop probability of 0.4. We apply temporal smoothing as a convolutional kernel over the temporal domain before the last classifier layer with a kernel size of $T$ (all frames).

\section{DFAUST Dataset extension}
We extend the DFAUST dataset for the task of action recognition. The DFAUST dataset \cite{dfaust:CVPR:2017} provides high-resolution 4D scans of human subjects in motion. It includes over 100 dynamic actions of 10 subjects (6:4 male-to-female ratio) with varying body shapes represented as registrations of aligned meshes. This dataset was not specifically designed for action understanding, however, it provides point cloud sequences with action labels per sequence. We extended it to our task by subdividing the dataset into clips of 64 frames of train and test human subjects. The trainingset is composed of three male and three female subjects and includes a total of 76 sequences, 395 clips, and $\sim$25K point cloud frames.  The testset is composed of two male and two female subjects and includes a total of 53 sequences, 313 clips, and $\sim$20K frames. This split was chosen in order to guarantee no subject will appear in both training and test splits as well as to make sure that all actions appear both in the train and test splits. Note that not all actions are performed by all subjects.

\begin{figure}[t]
    \centering
    \includegraphics[width=\linewidth]{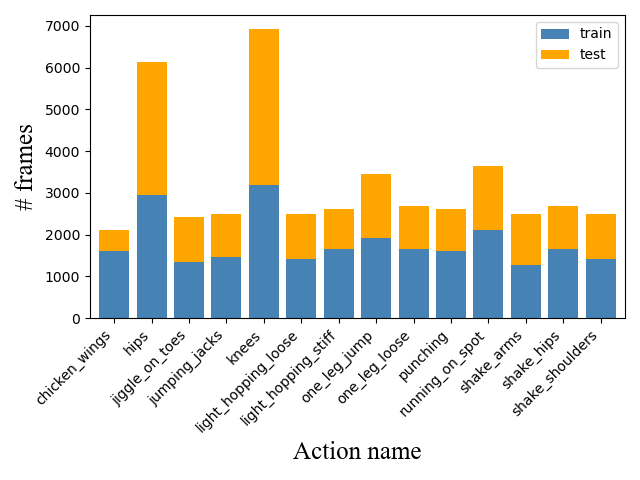}
    \caption{Our extension of the DFAUST dataset for action recognition. Number of frames per action in the training (blue) and test (orange) sets.}
    \label{fig:sup:dfaust_Stats}
\end{figure}

The action instance occurrences and full action list are depicted in \figref{fig:sup:dfaust_Stats}. It shows that there are two dominant classes (\textit{hips} and \textit{knees}). Therefore, to mitigate this imbalance's effect on training we use a weighted sampler that uses a sampling probability that is inversely proportional to a class's occurrence in the training set. 

\section{IKEA ASM preprocessing} The IKEA ASM is a large scale dataset and each frame may contain a very large number of points (hundreds of thousands). Most of the points, however, lie on static background regions. In order to keep the training time reasonable, we downsampled each frame using farthest point sampling (FPS) to have a fixed number of 4096 points. We then saved the data into clips for training. This reduced the training time for all methods significantly since the data loading was a bottleneck. 

\section{Additional Experiments}
\begin{figure}[t]
    \centering
    \includegraphics[width=\linewidth]{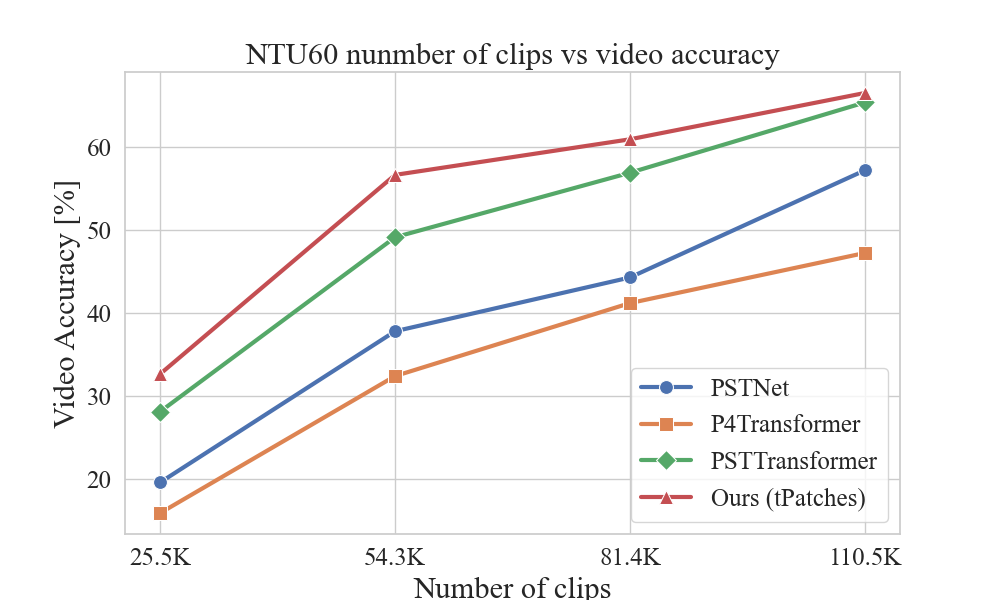}
    \caption{Video Accuracy as a function of the data fraction on the NTU 60 dataset. The results show that the proposed approach achieves better performance than PSTTransformer \cite{fan2021p4transformer}, P4Transformer \cite{fan2022psttransformer} when the dataset scale is limited.}
    \label{fig:ntu60_acc_datafrac}
\end{figure}

\textbf{Results on NTU RGB+D dataset.}
We conduct thorough experiments to evaluate the performance of our proposed approach compared to existing state-of-the-art methods on the NTU RGB+D 60 dataset \cite{shahroudy2016ntu}. Since the dataset is essentially saturated and previous methods show correlation between the different training splits (subject, view, and setup) we explore the performance as a function of the size of available data. Since the full NTU60 includes $57.6K$ videos and $\sim1.1M$ clips we evaluate the performance in terms of small fractions of the original data, specifically $(2.5\%, 5\%, 7.5\%, 10\%)$ that amount to  $(\sim25.5K, \sim54.3K, \sim81.4, \sim110.5K )$  clips respectively. The results are reported in \figref{fig:ntu60_acc_datafrac}. The results show that while some methods thrive when a lot of data is available, the proposed method demonstrates superiority even when the dataset scale is limited. For completeness, the performance of all methods when all data is available are comparable and given here: 90.2, 91.0, 89.3 for P4Transformer \cite{fan2021p4transformer}, PSTTransformer \cite{fan2022psttransformer} and the proposed approach respectively. 

\begin{figure}[t]
    \centering
    \includegraphics[width=\linewidth]{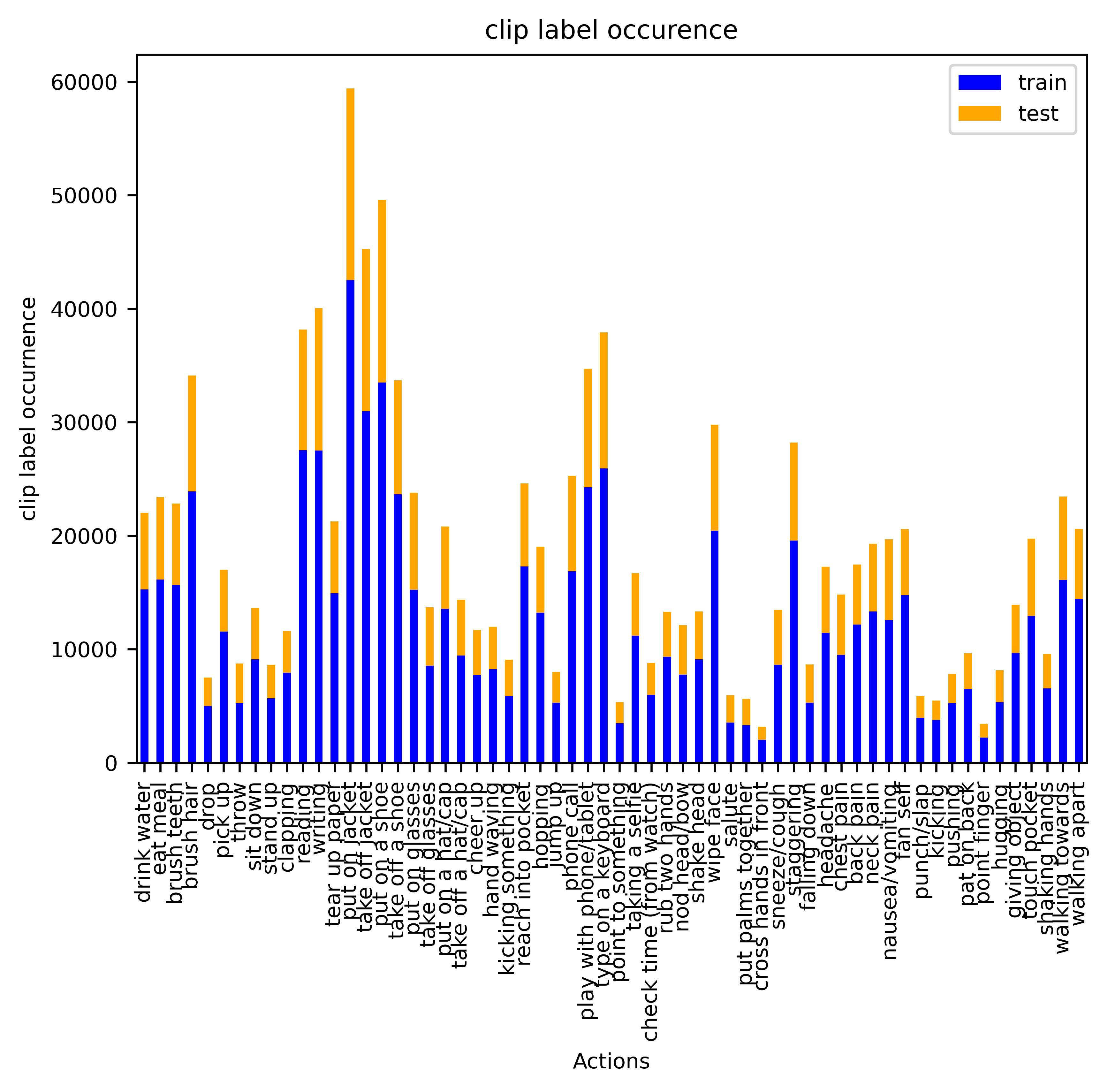}
    \caption{NTU60 clip class distribution. It shows an imbalance between different classes.}
    \label{fig:ntu60_class_imbalance}
\end{figure}

In this experiment we use the same training and test protocol as specified by P4Transformer \cite{fan2021p4transformer} and retrain the models using the parameters reported in the papers. Since the proposed approach and architecture were designed for per-frame prediction we add a 3 layer GRU at the end, to aggregate the temporal domain into a single vector representation per clip. To fairly compare the quality of the representation we also replace our classifier with the same classifier architecture as in P4Transformer and PSTTransformer. This helps avoid changes in performance that are related to the classifier and focus the evaluation on the core representation ability.  
To generate the data fractions, we uniformly sample a fraction of the videos in both train and test sets. This way, the distribution of class occurrences is preserved. Note that while the number of videos in each class is equal in both trainingset and testset, the lengths are different and therefore the number of generated clips is different. This causes a significant class imbalance, as shown in \figref{fig:ntu60_class_imbalance}.

\noindent\textbf{Absence of dynamic method's baselines on DFAUST and IKEA ASM.}
In \secref{Sec:results} we report the performance of various baseline methods on existing datasets for 3D action recognition. 
For DFAUST and IKEA ASM we report static methods PointNet~\cite{qi2017pointnet},   PointNet$^{++}$~\cite{qi2017pointnet++}, and Set Transformer~\cite{lee2019settransformer} by applying them on each point cloud frame individually. Additionally, we report temporal methods like PSTNet~\cite{fan2021pstnet} and also implemented a temporal smoothing version of each static method (PoinNet+TS, Pointnet$^{++}$+TS, and Set Transformer+TS respectively) by learning the weights of a convolutional layer over the temporal dimension. 
Some dynamic methods performance do not appear in the tables as those were not reported in the original papers. Since a code is publicly available for some, we corresponded with the authors of PSTNet~\cite{fan2021pstnet}, P4Transformer~\cite{fan2021p4transformer} and PST-Transformer~\cite{fan2022psttransformer} in order to get  recommendations for testing these methods on our datasets, however,  results were very poor  for~\cite{fan2021p4transformer, fan2022psttransformer} and therefore not reported to avoid unfair judgment on these methods.

\begin{figure}
    \centering
    \begin{subfigure}[b]{0.45\linewidth}
        \centering
        \includegraphics[width=\linewidth, trim={8cm, 2cm, 8cm, 2cm}, clip]{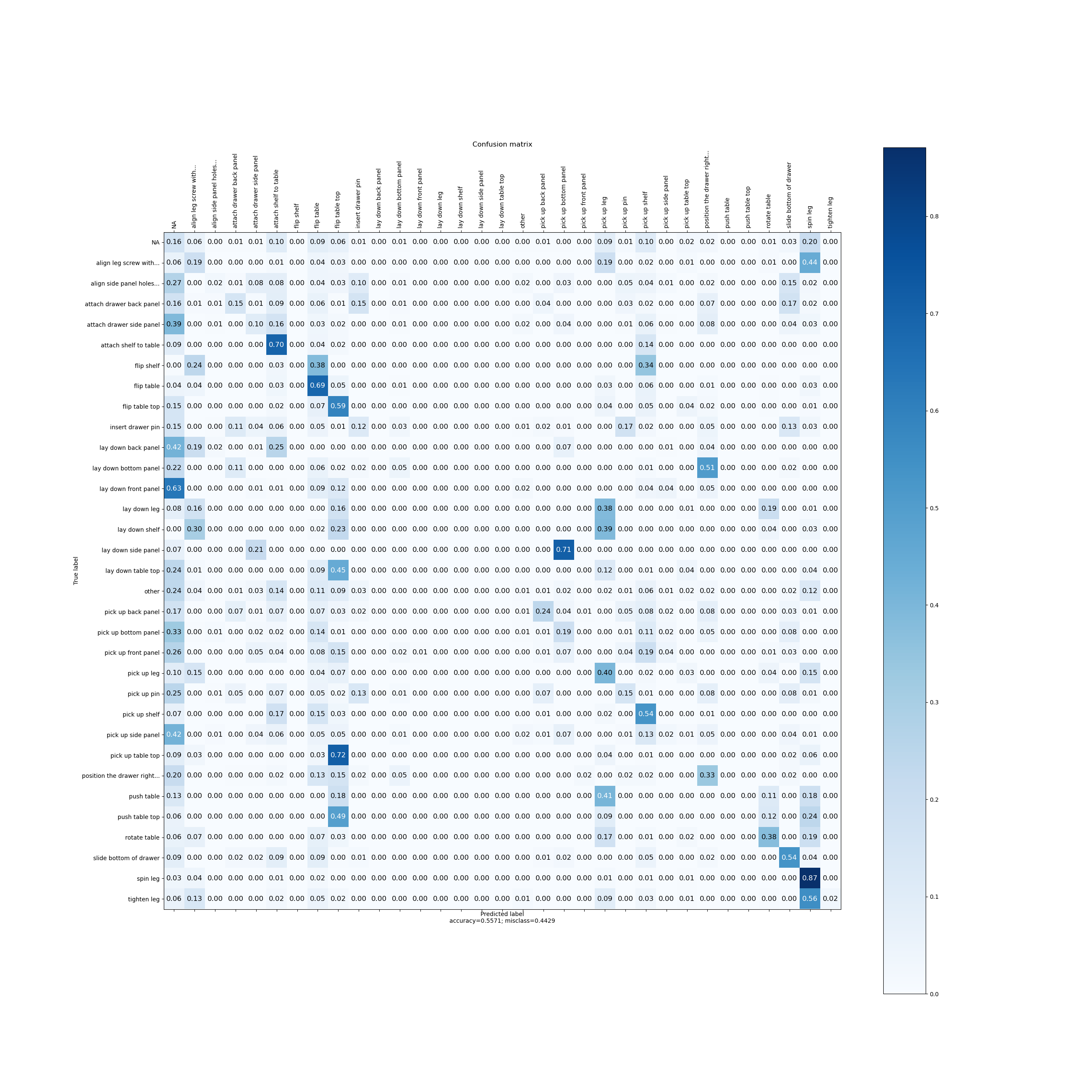}
    \end{subfigure}
    \hfill
    \begin{subfigure}[b]{0.45\linewidth}
        \centering
        \includegraphics[width=\linewidth, trim={8cm, 2cm, 8cm, 2cm}, clip]{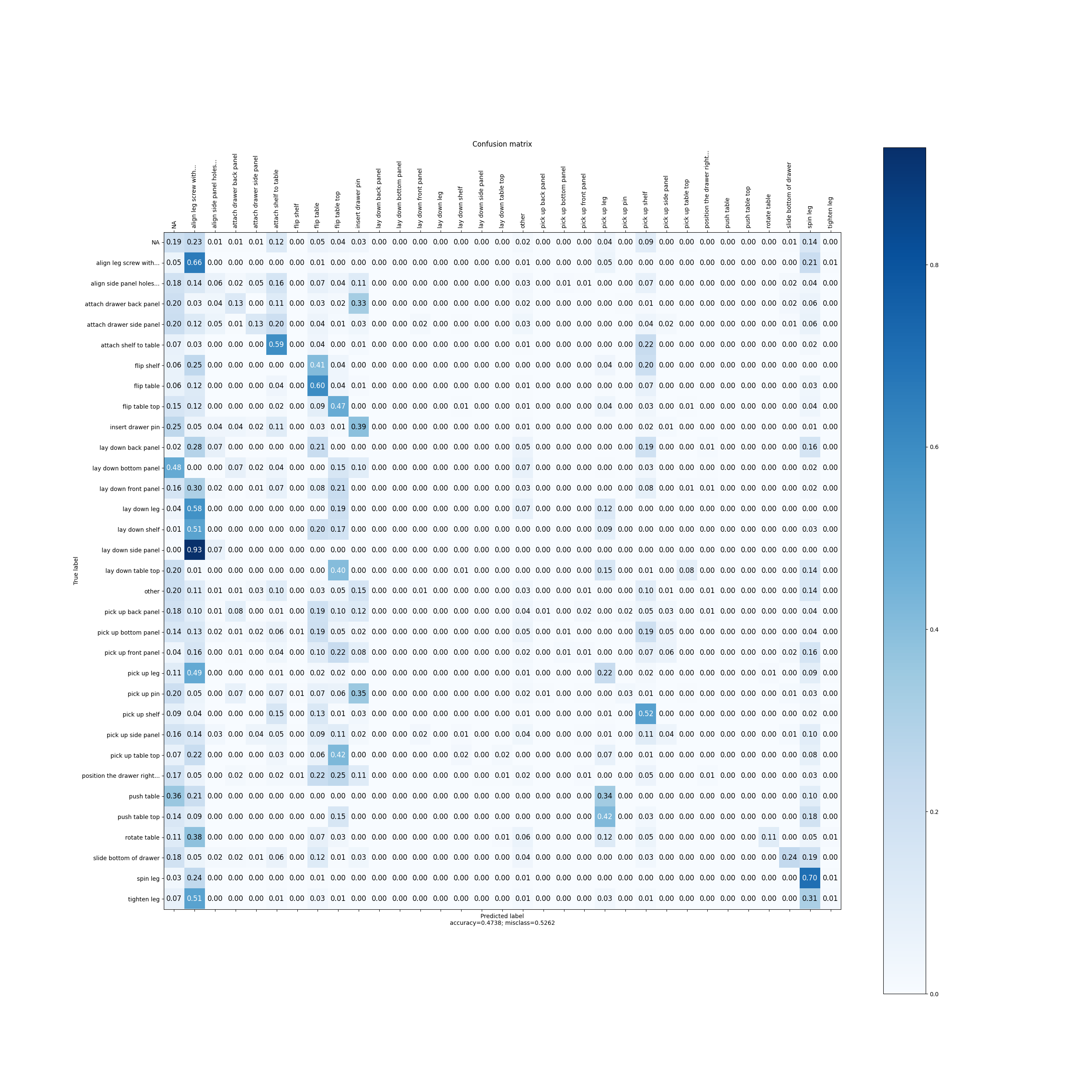}
    \end{subfigure}
    \caption{\textbf{Confusion matrices} for the proposed approach with (left) and without (right) bidirectional t-patches, evaluated on the IKEA ASM dataset.}
    \label{fig:results:confusion_matrices_ikea_asm}
\end{figure}

\noindent\textbf{IKEA ASM confusion matrices.} 
In \tabref{tab:results:baseline:action_segmentation_ikea_asm} of the main paper we presented quantitative results for action recognition on the IKEA ASM dataset \cite{ikeaasm}.Here, in \figref{fig:results:confusion_matrices_ikea_asm}, we present the confusion matrices for the proposed approach with and without bidirectional t-patches. Our analysis reveals varied impact on different action classes when Bi-directional t-patches (BD) are applied. Notably, the performance drops for classes like \textit{align leg screw} and \textit{align side panel}, while increasing for \textit{spin leg} and \textit{slide bottom drawer}. This variability stems from the data-dependence of BD, addressing temporal collapse, which can have a positive and negative effects by reintroducing informative and uninformative patches.

\begin{figure}[t]
        \centering
        \begin{subfigure}{.32\linewidth}
        \centering
        \includegraphics[width=0.99\linewidth, trim=300 65 300 100, clip]{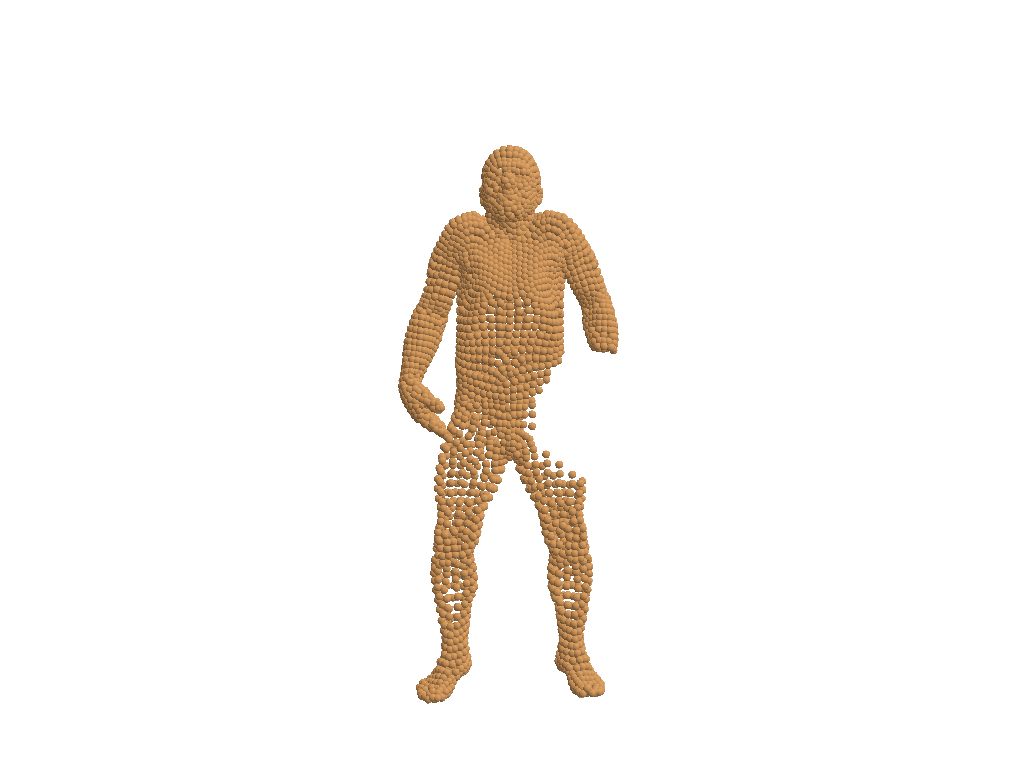}
        \end{subfigure}
                \begin{subfigure}{.32\linewidth}
        \centering
        \includegraphics[width=0.99\linewidth, trim=300 65 300 100, clip]{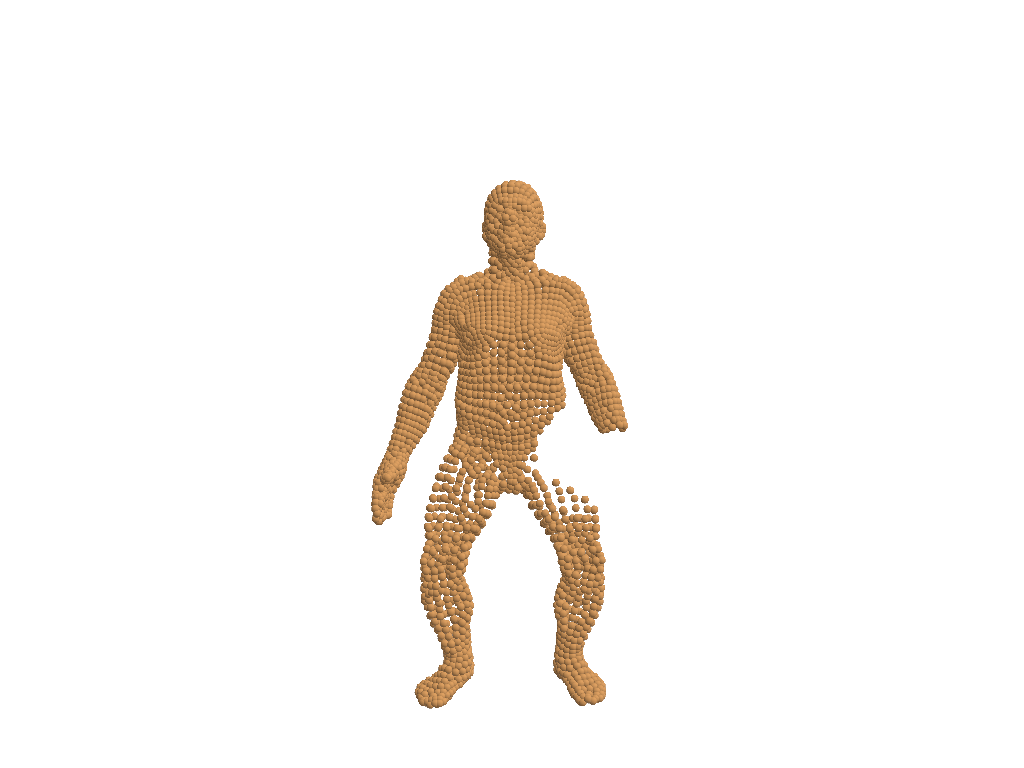}
        \end{subfigure}
                \begin{subfigure}{.32\linewidth}
        \centering
        \includegraphics[width=0.99\linewidth, trim=300 65 300 100, clip]{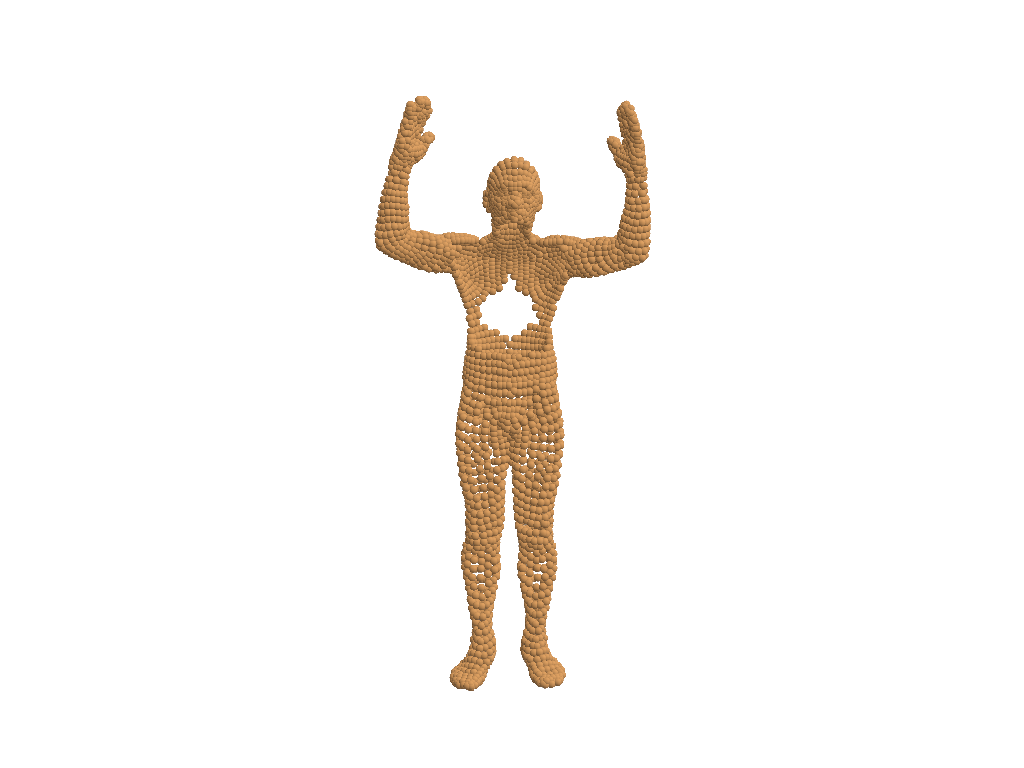}
        \end{subfigure}
        \caption{\textbf{Occlusion ablation}. Visualizing the point clouds used in the occlusion ablation experiment on the DFAUST dataset.}
        \label{fig:sup:occlusions}
\end{figure}

\noindent\textbf{Occlusion ablation study.}
We conducted additional experiments to evaluate the effectiveness of the bidirectional t-patches in the context of occlusions. The results, reported here in \tabref{tab:sup:occlusions} and visualized in \figref{fig:sup:occlusions}, show that the proposed approach provides a boost in performance which significantly benefits from the bidirectional sampling strategy. In this experiment, we simulated occlusions on the DFAUST dataset by randomly selecting a point from a random frame and removing all points within a $20\%$ radius for $16$ consecutive frames ($64$ frames per clip). 
\begin{figure}[tb]
    \centering
        \centering
         \begin{tabular}{l c c c}
         \toprule
             &   \multicolumn{2}{c}{\textbf{Frame acc.}} &    \\
            \textbf{Method}  & \textbf{top 1} & \textbf{top 3} &\textbf{mAP}\\
            \hline
            PointNet$^{++}$ + TS  &  66.71 & 86.80  & 0.7985\\  
            Ours &  76.75  & 95.54 &  \textbf{0.8770}\\
            Ours + BD &  \textbf{84.81}  & \textbf{96.66}  & 0.8683\\ 
            \hline
            Ours+BD no occlusion & 87.26 &99.26 &0.8616 \\
         \bottomrule
        \end{tabular}
    \caption{\textbf{Occlusion ablation} on DFAUST. Exploring effects of occlusions with and without bidirectional t-patches. For reference we include our ``no occlusion'' result from the main paper.}
    \label{tab:sup:occlusions}
\end{figure}

\noindent\textbf{Additional details on accuracy and mAP inconsistency.} 
Our models are trained using the cross entropy loss, which encourages high recall (i.e., Top 1 accuracy). The mAP metric roughly summarizes the trade-off between precision and recall and is important for cases where a clip contains multiple action labels. The reason for the small inconsistency between Top 1 and mAP is class imbalance.
In the IKEA ASM dataset, there is a significant imbalance with many occurrences for classes such as ``spin leg'' (which appears four times in each of the three furniture types).  Methods that perform well on dominant classes will be favored by the Top 1 metric. The DFAUST dataset is also affected by class imbalance, albeit less so than IKEA ASM. Importantly, when ranking by either Top 1 or mAP, our t-patch based method outperforms previous approaches.

\begin{figure*}
    \centering
    
    \begin{subfigure}{.24\linewidth}
    \centering
        \includegraphics[width=.99\linewidth]{assets/figs/ASM_vis/flip_table_tpatches_new/flip_table_0_img.png}
    \end{subfigure}
    \begin{subfigure}{.24\linewidth}
    \centering
        \includegraphics[width=.99\linewidth]{assets/figs/ASM_vis/flip_table_tpatches_new/flip_table_1_img.png}
    \end{subfigure}
        \begin{subfigure}{.24\linewidth}
    \centering
        \includegraphics[width=.99\linewidth]{assets/figs/ASM_vis/flip_table_tpatches_new/flip_table_2_img.png}
    \end{subfigure}
    \begin{subfigure}{.24\linewidth}
    \centering
        \includegraphics[width=.99\linewidth]{assets/figs/ASM_vis/flip_table_tpatches_new/flip_table_3_img.png}
    \end{subfigure}

        \begin{subfigure}{.24\linewidth}
    \centering
       \includegraphics[width=.99\linewidth, trim={7cm 5cm 3cm 0cm}, clip]{assets/figs/ASM_vis/flip_table_tpatches_new/flip_table_0_pc_bw.png}
    \end{subfigure}
    \begin{subfigure}{.24\linewidth}
    \centering
        \includegraphics[width=.99\linewidth, trim={7cm 5cm 3cm 0cm}, clip]{assets/figs/ASM_vis/flip_table_tpatches_new/flip_table_1_pc_bw.png}
    \end{subfigure}
        \begin{subfigure}{.24\linewidth}
    \centering
        \includegraphics[width=.99\linewidth, trim={7cm 5cm 3cm 0cm}, clip]{assets/figs/ASM_vis/flip_table_tpatches_new/flip_table_2_pc_bw.png}
    \end{subfigure}
    \begin{subfigure}{.24\linewidth}
    \centering
        \includegraphics[width=.99\linewidth, trim={7cm 5cm 3cm 0cm}, clip]{assets/figs/ASM_vis/flip_table_tpatches_new/flip_table_3_pc_bw.png}
    \end{subfigure}

            \begin{subfigure}{.24\linewidth}
    \centering
       \includegraphics[width=.99\linewidth, trim={7cm 5cm 3cm 0cm}, clip]{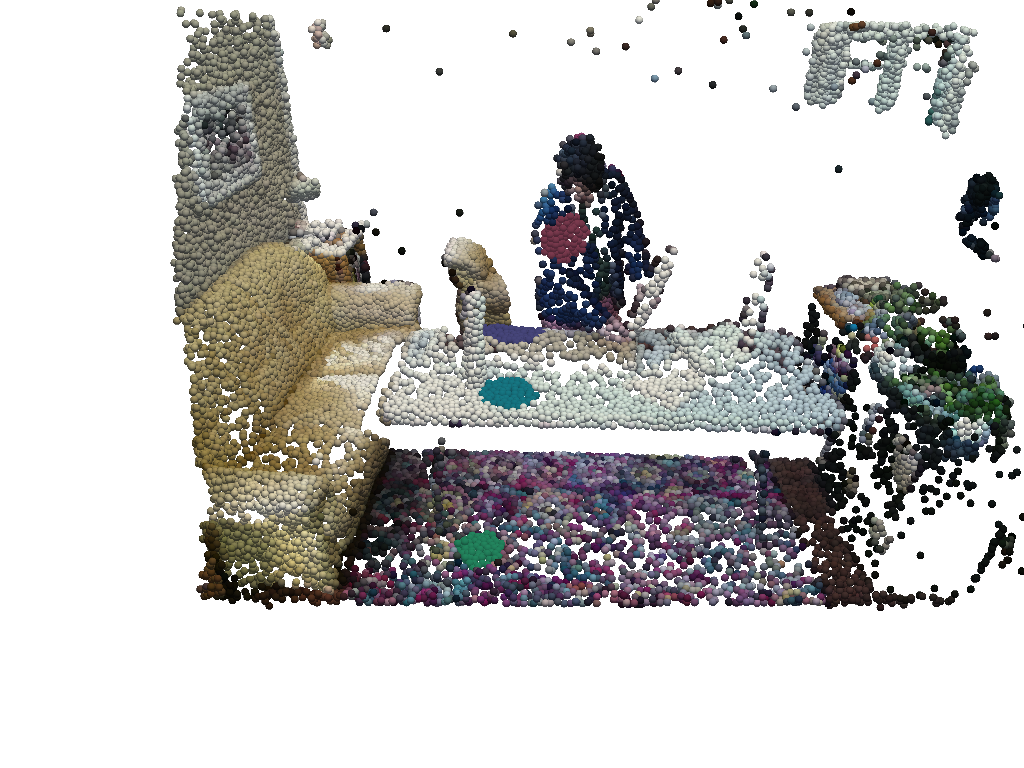}
    \end{subfigure}
    \begin{subfigure}{.24\linewidth}
    \centering
        \includegraphics[width=.99\linewidth, trim={7cm 5cm 3cm 0cm}, clip]{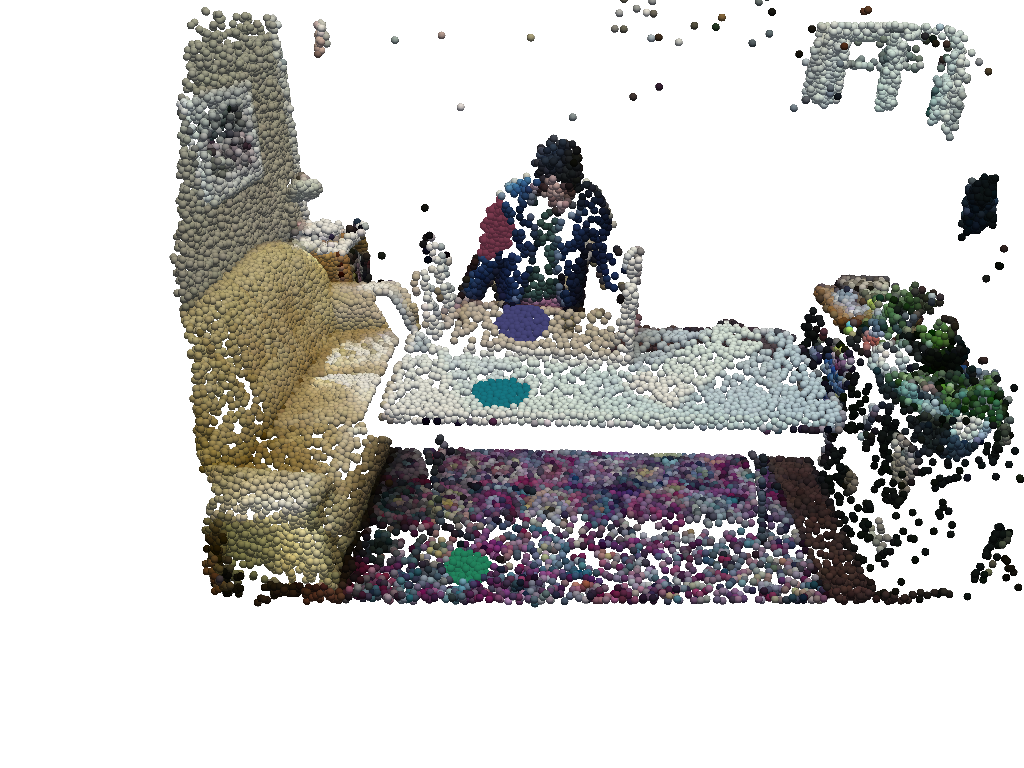}
    \end{subfigure}
        \begin{subfigure}{.24\linewidth}
    \centering
        \includegraphics[width=.99\linewidth, trim={7cm 5cm 3cm 0cm}, clip]{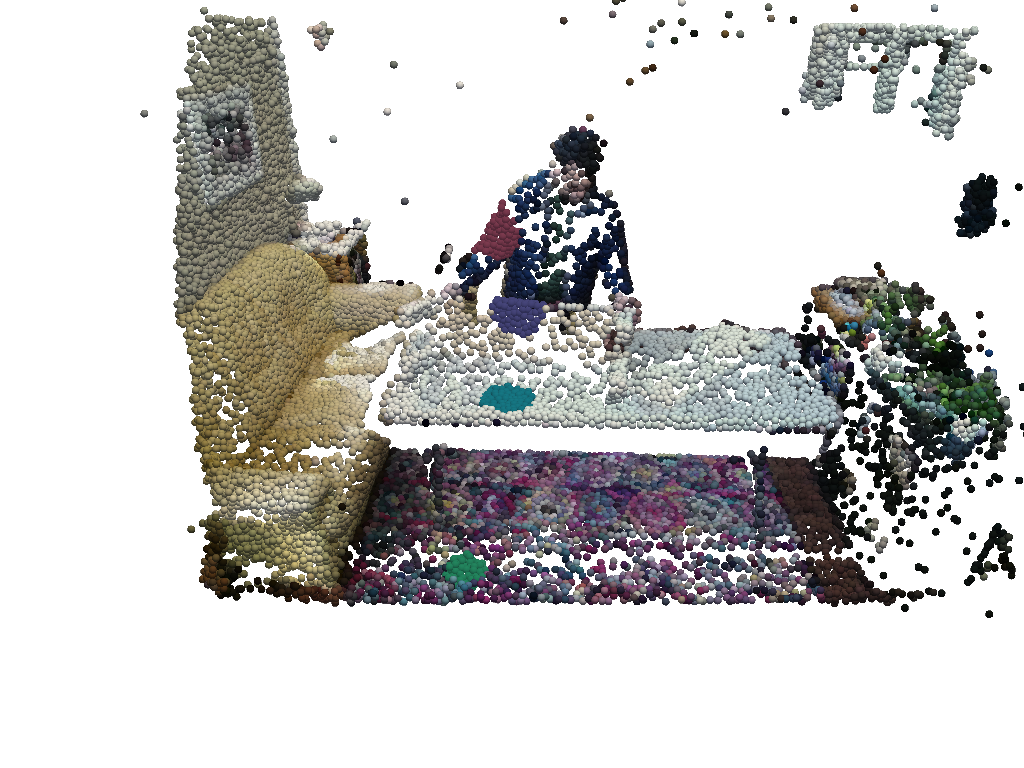}
    \end{subfigure}
    \begin{subfigure}{.24\linewidth}
    \centering
        \includegraphics[width=.99\linewidth, trim={7cm 5cm 3cm 0cm}, clip]{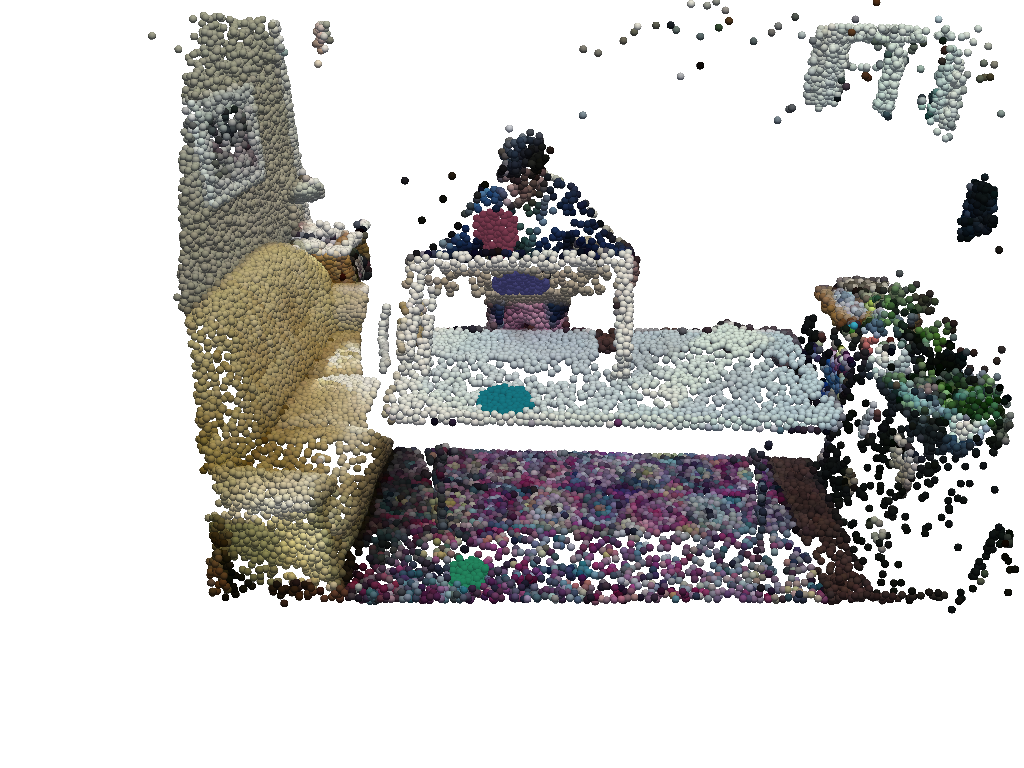}
    \end{subfigure}
    \caption{\textbf{IKEA ASM example with t-patches.}  The flip table action for the TV Bench assembly is visualization including the RGB image (top), a grayscale 3D point cloud with colored t-patches (middle), and a RGB colored 3D point cloud with colored t-patches (bottom). t-patches are highlighted in color. The \textbf{\textcolor{tpatch_blue}{blue}} is on the moving TV Bench assembly, \textbf{\textcolor{tpatch_pink}{maroon}} is on the moving persons arm,  \textbf{\textcolor{tpatch_cyan}{teal}} is on the static table surface, and \textbf{\textcolor{tpatch_green}{green}} is on the colorful static carpet.}
    \label{fig:IKEA_ASM_flip_table_full}
\end{figure*}

\noindent\textbf{Additional IKEA ASM Visualization.} 
In \figref{fig:IKEA_ASM_flip_table} we provided a visualization of the t-patches with grayscale point clouds. Here, in \figref{fig:IKEA_ASM_flip_table_full} we provide an extended version of that figure that also includes colored point clouds. 

\noindent\textbf{Parameter ablation for t-patch extraction. }
In \secref{sec:ablation_study} we performed an ablation study and analyzed the effect of tuning the number of neighbors to extract ($k$) and the number of points to sub sample ($n$) on accuracy performance. The results were reported in \tabref{tab:ablation:k_n}. Here we complement this ablation by also reporting the time and number of network parameters for each parameter selection in \tabref{tab:appx:ablation:k_n}. Timing experiment was done on an NVIDIA A5000 GPU (timing experiment for \figref{tab:time_memory_sota} was done on an A100, hence the reported time differences).   

\begin{table}[] 
    \centering 
    \setlength\tabcolsep{4pt}
    \begin{tabular}{c c c c c c}
         \toprule
            &&  \multicolumn{2}{c}{\textbf{Frame acc.}}    \\
            \textbf{n} & \textbf{k} & \textbf{top 1} & \textbf{top 3} &  \textbf{mAP} & \textbf{Time [ms]} \\
            \hline
            256 & 16 & 76.96 & 97.54 & 0.8430 &385\\
            512 & 16 & \textbf{80.03} & \underline{97.57} & \textbf{0.8975} &415\\
            1024 & 16 & 77.30 & \textbf{97.88} &  \underline{0.8507}&448\\
            512 & 8 & 76.87 & 96.21  & 0.7557 &324\\
            512 & 32 & \underline{77.91} & 96.60 & 0.7453 & 608\\
         \bottomrule
    \end{tabular}
    \caption{\textbf{t-patch parameters ablation}. Results for the number of neighboring points in a patch $k$ and number of downsampled points $n$ show that the method is robust.}
    \label{tab:appx:ablation:k_n}
\end{table}

\end{document}